\RequirePackage{fix-cm}
\documentclass[smallextended]{svjour3}       % onecolumn (second format)
\smartqed  % flush right qed marks, e.g. at end of proof
\usepackage{graphicx}
\usepackage{amsmath,amssymb} % define this before the line numbering.
\usepackage{color}
\usepackage{times}
\usepackage{epsfig}
\usepackage{graphicx}
\usepackage{amsmath}
\usepackage{amssymb}
\usepackage{textcomp}
\usepackage{float}
\restylefloat{figure}
\usepackage{subfig}
\usepackage{url}
\usepackage{verbatim}
\usepackage{breqn}
\usepackage{enumerate}
\floatstyle{plaintop}
\restylefloat{table}
\usepackage{paralist}
\usepackage{natbib}
\usepackage{tikz}
\usepackage{breqn}
\usepackage{hyperref}
\hypersetup{
     colorlinks=true,citecolor=blue,pagebackref=true,breaklinks=true,letterpaper=true,bookmarks=false
}

\DeclareMathOperator{\dis}{d}

\graphicspath{{./Resources/}}
\DeclareGraphicsExtensions{.pdf,.jpeg,.jpg,.png}

\begin{document}

\title{Sketch-based Image Retrieval from Millions of Images under Rotation, Translation and Scale Variations%\thanks{Grants or other notes
%about the article that should go on the front page should be
%placed here. General acknowledgments should be placed at the end of the article.}
}
% \subtitle{Do you have a subtitle?\\ If so, write it here}

\titlerunning{Sketch-based Image Retrieval from Millions of Images}        % if too long for running head

\author{Sarthak Parui         \and
        Anurag Mittal %etc.
}

%\authorrunning{Short form of author list} % if too long for running head

\institute{Sarthak Parui \and Anurag Mittal \at
              Computer Vision Lab, \\Dept. of Computer Science \& Engineering \\Indian Institute of Technology Madras\\
              Tel.: +91-44-22575352\\
              \email{{sarthak,amittal}@cse.iitm.ac.in}           %  \\
%             \emph{Present address:} of F. Author  %  if needed
}

\date{Received: date / Accepted: date}
% The correct dates will be entered by the editor

\maketitle

\begin{abstract}
Proliferation of touch-based devices has made sketch-based image retrieval practical. 
While many methods exist for sketch-based object detection/image retrieval on small datasets, relatively less work has been done on large (web)-scale image retrieval. In this paper, we present an efficient approach for image retrieval from millions of images based on user-drawn sketches. 
Unlike existing methods for this problem which are sensitive to even translation or scale variations, our method handles rotation, translation, scale (i.e. a similarity transformation) and small deformations.
The object boundaries are represented as chains of connected segments and the database images are pre-processed to obtain such chains that have a high chance of containing the object. This is accomplished using two approaches in this work: a) extracting long chains in contour segment networks and b) extracting boundaries of segmented object proposals.
\iffalse For efficient online processing, each database image is preprocessed to extract long sequences of contour segments (chains) using two complimentary methods. \fi
These chains are then represented by similarity-invariant variable length descriptors. 
Descriptor similarities are computed by a fast Dynamic Programming-based partial matching algorithm.
This matching mechanism is used to generate a hierarchical k-medoids based indexing structure for the extracted chains of all database images in an offline process which is used to efficiently retrieve a small set of possible matched images for query chains. 
% which is integrated in a hierarchical k-medoids based indexing structure for very fast online retrieval.
Finally, a geometric verification step is employed to test geometric consistency of multiple chain matches to improve results. 
Qualitative and quantitative results clearly demonstrate superiority of the approach over existing methods.

\end{abstract}

\keywords{Sketch-based Retrieval \and Image Retrieval \and Shape Representation \and Indexing}
%-------------------------------------------------------------------INTRODUCTION ------------------------------------------------------------------------------------

\section{Introduction}

%########################################################################################### Motivation ####################################

The explosive growth of digital images on the web has substantially increased the need for an accurate, efficient and user-friendly large-scale image retrieval system. 
With the growing popularity of touch-based smart computing devices and the consequent ease and simplicity of querying images via hand-drawn sketches on touch screens~\citep{lee2011shadowdraw}, sketch-based image retrieval has emerged as an interesting application.
The standard mechanism of text-based querying could be imprecise due to wide demographic variations. It also faces the issue of availability, authenticity and ambiguity in the tag and text information surrounding an image~\citep{sigurbjornsson2008flickr,schroff2011harvesting}, which necessitates us to exploit image content for better search. 
Although various popular image search engines such as Google\footnote{\url{https://images.google.com/}} and TinEye\footnote{\url{https://www.tineye.com/}} provide an interface for similar image search using an exemplar image as the query, a user may not have access to such an  image every time. Instead, a hand-drawn sketch may be used for querying since sketching is a fundamental mechanism for humans to conceptualize and render visual content as also suggested by various Neuroscience studies~\citep{marr1982computational,landay2001sketching,walther2011simple}. 
Thus, Sketch-based image retrieval, being a far more expressive way of image search, either alone or in conjunction with other retrieval mechanisms such as text, may yield better results, which makes it an important and interesting problem to study. 

Apart from online web-scale image retrieval, an efficient sketch-based image retrieval mechanism has numerous other applications. For instance, it can be used to efficiently retrieve intended images from a constrained dataset, viz. a user's personal photo-album in a touch-sensitive camera/tablet for which no text/tag information is available. Observing the expressive power of free-hand user sketches, a few methods have been proposed for searching/designing apparels~\citep{king1996feature,tseng2009efficient,zhang2012apparel,kondo2014sketch}, accessories~\citep{zeng2014sketch2jewelry} or generic 3D objects such as home appliances~\citep{eitz2012sbsr} using user-sketches. These applications require efficient shape representation and fast sketch-to-image matching to facilitate a smooth user experience. Furthermore, sketch-based retrieval can also be used to improve existing text-based image retrieval systems. 
For instance, it may be possible to build a sketch in an on-line manner using the first few results of a text query system~\citep{lee2009shape,bagon2010detecting,marvaniya2012drawing} and use this sketch for retrieving images that may not have any associated tag information. Image tag information may also be improved for a database in an off-line process using sketch-based retrieval.

%########################################################################################### End of Motivation ####################################

%########################################################################################### Shape-based Object Detection in small scale ####################################
\paragraph{Sketch-based Object Detection: }
Several approaches have been considered in the literature for describing shapes and measuring their similarity.
\iffalse To represent the shape structure of the objects present in an image, typically edge-detection/boundary detection is performed as a pre-processing task and the location and orientation information contained in the edge pixels (edgels) are utilized for determining shape correspondence. \fi
Basic approaches for measuring the similarity of rigid shapes include Chamfer Matching~\citep{barrow1977parametric}, \iffalse ~\cite{barrow1977parametric} proposed Chamfer Matching \fi in which the binary template of a target shape is efficiently matched in an edge-image by calculating the distances to the nearest edge pixels. This process can be sped up using a pre-computed Distance Transform of such an edge-image.\iffalse a priori for each pixel in the image (Distance Transform), which helps in capturing the possible geometric transformations of the reference template for efficient edgemap alignment.\fi~\cite{huttenlocher1993comparing} used a closely related point correspondence measure called the Hausdorff distance which examines the fraction of points in one set that lie within some $\varepsilon$ distance of points in the other set (and vice versa).~\cite{liu2010fast} improved the accuracy and efficiency of edgemap alignment substantially by incorporating image edge-orientation information and using a three-dimensional distance transform to match over possible locations and orientations.

To handle non-rigidity of shapes and speed up matching,~\cite{belongie2002shape} proposed \emph{Shape Context} in which a shape is described by a set of descriptors, each of which captures the spatial distributions of other points along the shape contour in a log-polar space.~\cite{ling2007shape} extend \emph{Shape Context} by using the \emph{inner-distance} instead of the Euclidean distance which restricts the paths between any two contour points to remain within the shape, thus making the descriptors quite robust to articulations.~\cite{gopalan2010articulation} further normalize each part affinely before computing the innner distance for achieving even greater invariance to shape variations, especially in the case of perspective distortions. 
Shape Context-based methods have shown a good promise for clutter free shape-to-shape matchings. However, for matching shapes in images where an image has lots of extra and/or clutter edges, these are difficult to apply.

Many methods have been proposed in the literature for more sophisticated sketch-based Object Detection/Retrieval in images that handle more shape variations, although the running time for detection and retrieval is often compromised.~\cite{felzenszwalb2007hierarchical} use a \emph{shape-tree} to form a hierarchical structure of contour segments for representing each object, which helps in capturing local geometric properties along with the global shape structure. Deformation is allowed at individual nodes of the tree and an efficient Dynamic Programming-based matching algorithm is used to match two curves.\iffalse In order to address the issues of background clutter and intra-class shape variations,\fi~\cite{ferrari2006object} create a Contour Segment Network (CSN) by connecting nearby edge pixels in an edge-map. For matching a sketch to an image, they find paths through the constructed CSN that resemble the outline of the sketch. In a later approach,~\cite{ferrari2008groups} propose groups of \emph{k} (typically $\leq 4$) Adjacent approximately straight Segments (\emph{k}AS) as a local feature and describe them in a scale insensitive way. These \emph{k}ASs are extracted for a large number of image windows and finally the object boundary is traced by linking individual small matched \emph{k}ASs in a multi-scale detection phase. In further work, they learn a codebook of Pairs of Adjacent Segments (PAS)~\citep{ferrari2010images} which is used in combination with Hough-based centroid voting and non-rigid thin-plate spline matching for detecting the sketched object in cluttered images. 
In contrast to \emph{k}AS matching,~\cite{ravishankar2008multi} propose a multi-stage contour-based detection approach, where Dynamic Programming is used to match segments directly to the edge pixels.

Recently, several approaches have shown a good Object Detection performance using self-contained angle-based descriptors for representing the object contours.~\cite{lu2009shape} propose a shape descriptor based on a three-dimensional histogram of angles and distances for triples of consecutive sample points along the object contours. To explicitly handle varying local shape distortions, they exploit a particle filter framework to jointly solve the contour fragment grouping and matching problems. 
\iffalse Donoser \emph{et al.}~\cite{donoser2010efficient} analyze angles between any two points and a fixed third point on a close contour and design an \emph{integral image}-based mechanism to match whole object boundary. Similar to~\cite{donoser2010efficient}, \fi~\cite{riemenschneider2010using} sample a contour into a fixed number of points and calculate the angles between a line connecting any two sampled points and a line to a third point relative to the position of the first two points. This representation is insensitive to translation and rotation but not scale. They employ a partial matching mechanism between two such angle-based descriptors by efficiently choosing the range of consecutive points using an integral image-based approach. To achieve robustness with respect to scale, object detection is performed over a range of scales. 
%They employ a partial matching mechanism and a verification stage for detecting objects for a given sketch. 

All of these methods and many other state-of-the-art methods for Object Detection and Retrieval~\citep{scott2006robust,kokkinos2011inference,ma2011partial,yarlagadda2012meaningful} \iffalse typically represent objects using small contour fragments and trace the object boundary by linking them at a costly multi-scale detection phase. Furthermore, they \fi employ expensive online matching operations based on complex shape features to enhance the detection performance and typically show results on relatively small-sized datasets such as ETHZ~\citep{ferrari2006object} and MPEG-7~\citep{latecki2000shape} containing only a few hundreds or thousands of images while taking a considerable time to parse through each image at detection/search time. However, for a dataset with millions of images with a desired retrieval time of at most a few seconds, these methods are inapplicable/insufficient. Efficient Image pre-processing and a mechanism for fast online retrieval are necessary for large (web)-scale Image Retrieval.

%########################################################################################### End of Shape-based Object Detection in small scale ####################################

%########################################################################################### LSSBIR literatures ####################################
\paragraph{Large-Scale Image Retrieval using Sketches: }
% Relatively fewer attempts have been made in the literature on the problem of sketch-based image retrieval on large databases. 
Only a few attempts exist in the literature for the problem of sketch-based image retrieval on large databases.~\cite{eitz2009descriptor} decompose an image or sketch into different spatial regions and measure the correlation between the direction of strokes in the sketch and the direction of gradients in the image by proposing two types of descriptors, viz. an Edge Histogram Descriptor and a Tensor Descriptor. Histograms of prominent gradient orientations are encoded in the Edge Histogram Descriptor, whereas the Tensor Descriptor determines a single representative vector per cell that captures the main orientation of the image gradients of that cell. Descriptors at corresponding positions in the sketch and the image are correlated for matching.
\iffalse They assume that the descriptors for individual spatial regions relative to the overall image will have a similar footprint if an image is similar to a sketch. \fi
Due to this strong spatial assumption, they fail to retrieve images if the sketched object is present at a different scale, orientation and/or position. To determine similar images corresponding to a sketch, a linear scan over all database images is performed. This further limits the scalability of the 
method for large databases.
% where a normalized frequency histogram representing the distribution of GF-HOG derived ``visual words'' is constructed. 
% These descriptors are encoded into Bag-of-Visual-Words and a normalized frequency histogram representing the distribution of descriptor ``words'' is constructed.
% Hu \emph{et al.}~\cite{hu2010gradient} employ a similar \emph{gradient vector field}-based approach to encapsulate the spatial structure of gradient images and use the Bag-of-Words model to facilitate retrieval. However, since the experiments were performed only on small datasets that contain less than $400$ images, the usability of the method on large scale retrieval is not very clear.

To address the issue of scalability,~\cite{cao2011edgel} propose an indexing-friendly raw contour based method.  Given a sketch query, the primary objective of this method is to retrieve database images which closely correlate with the shape and the position of the sketched object. For every possible image location and a few orientations, they generate an inverted list of images that have edge pixels (edgels) at that particular location and orientation. For a sketch query, similar images are determined by counting the number of similar edgels in both the sketch and images, which makes this method susceptible to scale, translation and rotation changes.~\cite{bozas2012large} introduce a hashing based framework with a strong assumption that a user only wants spatially consistent images as the search result. They extract HoG features~\citep{dalal2005histograms} for overlapping spatial patches in an image and represent them using binary vectors by thresholding the HoG responses. The similarity between corresponding patches in the sketch and the image is estimated using the \emph{Min-hash} algorithm~\citep{chum2008near} that exploits the set-overlap similarity of these binarized descriptors.
\iffalse They generate HoG features~\cite{dalal2005histograms} for overlapping spatial patches in an image and the similarity between corresponding patches in the sketch and in the image is estimated using \emph{Min-hash} algorithm. \fi
Similar to~\cite{cao2011edgel}, a reverse indexing structure on the hash keys is built to facilitate fast retrieval. 

In order to encapsulate the spatial structure,~\cite{hu2010gradient} describe both images and sketches using \emph{Gradient field HoG} (GF-HOG) which encodes a sparse orientation field computed from the gradients of the edge pixels. To facilitate retrieval, a Bag of Visual Words model is used and sketch-to-image similarity is measured by computing the distance between corresponding frequency histograms representing the distribution of GF-HOG derived ``visual words''. However, this representation is noisy in presence of even small amount of background clutter. Moreover, since the experiments were performed only on small datasets that contain less than $400$ images, the usability of the method on large scale retrieval is not very clear. 

~\cite{riemenschneider2011image} extend their prior idea~\citep{riemenschneider2010using} to large scale retrieval by building a \emph{vocabulary tree}~\citep{nister2006scalable} on the descriptors. 
They extract heavily overlapping contour fragments (allowing only one point shift) from a sketch and the edge-map of an image. In their approach, each contour is composed of a fixed number of points $L$ and described as a matrix of $\binom{L}{2}$ angles that denote the orientation of the lines joining any two such points with respect to a vertical line. This makes their method sensitive to scale and orientation changes. Furthermore, due to the use of dense descriptors, the computational complexity is still very high for large datasets. Moreover, the experiments were performed only on very small datasets consisting of less than $300$ images and therefore the applicability of this method in large scale is again not very clear.
%  The authors also did not perform any experiments on a large dataset. 
\iffalse To alleviate the issue of scale, translation and rotation sensitivity, \fi ~\cite{zhou2012sketch} determine the most ``salient'' object in the image and measure image similarity based on a descriptor built on the object. However, determining saliency is a very hard problem and the accuracy of even the state-of-the-art saliency methods in natural images is low~\citep{li2014secrets}, thus rendering the method possibly quite unreliable. \iffalse which makes this method possibly quite unreliable. \fi

%########################################################################################### End LSSBIR literatures ####################################

%########################################################################################### Our Method in Context ####################################
\paragraph{Proposed method in context: }

In this paper, we develop a system for large scale sketch-based image retrieval that can handle scale, translation and rotation (similarity) variations without compromising on efficiency, which we believe has not been addressed earlier in the literature. 
First, the essential shape information of all the database images is captured by extracting sequences/chains of contour segments in an offline process using two efficient and often complimentary methods: (a) finding long connected segments in contour segment networks (Sec.~\ref{sec:csn}) and (b) using boundaries of segmented object proposals (Sec.~\ref{sec:gop}).
Such chains are represented using a similarity-invariant variable length descriptors (Sec.~\ref{sec:descriptor}). These chain descriptors are matched using an efficient Dynamic Programming-based \emph{approximate substring} matching algorithm. Note that, variability in the length of the descriptors makes the formulation unique and more challenging. Furthermore, partial matching is allowed to accommodate intra-class variations, small occlusions and the presence of non-object portions in the chains (Sec.~\ref{sec:matching_two_chains}).
A hierarchical indexing tree structure of the chain descriptors of the entire image database is built offline to facilitate fast online search by matching the chains down the tree (Sec.~\ref{sec:image_indexing_and_fast_retrieval}). 
Finally, a geometric verification scheme is used for refining the retrievals by considering the geometric consistency among multiple chain matchings (Sec.~\ref{sec:image_retrieval}). 
Results on several datasets indicate superior performance and advantages of our approach compared to prior work \iffalse, along with its strengths and weaknesses\fi (Sec.~\ref{sec:experiments}).

\section{From Images to Contour Chains}
\label{sec:image_representation}
In this section, we describe offline preprocessing of database images with an objective of having a compact representation which can be used to efficiently match the images with a query sketch. We first note that a user typically draws an object along its boundary~\citep{cole2008people} and a sketch of the object boundaries can more or less capture the distinctive object shape information~\citep{cole2009well}. Thus, an image representation based on contour information of the object boundaries would be quite appropriate in this scenario.~\cite{ferrari2008groups} construct an unweighted \emph{Contour Segment Network} (CSN) which links nearby edge segments, and then extract $k(\leq4)$ adjacent segments($k$AS) from such a network.  Full contour segment networks~\citep{ferrari2008groups} capture a good amount of information about the image edge segments, but are difficult to represent compactly and match efficiently in a large database. Shape Descriptors can be utilized; but they can be quite noisy, especially in the presence of image clutter.  In this work, we represent shape information using (long) chains of contour segments which we believe contain a good amount of information for capturing the shape perspective and are efficient for storage and matching at the same time.  While extraction of contour chains in sketches is trivial, doing so in database images is non-trivial due to the exponential number of chain possibilities.  In this work, we devise two efficient and often complimentary methods for extracting and encoding the essential object boundary information in a database image by (a) finding long chains in contour segment networks, and (b) using boundaries of segmented object proposals.

%-------------------------------------------------Subsection: Finding Long Chains in Contour Segment Network --------------------------------------------
\subsection{Finding Long Chains in Contour Segment Network}
\label{sec:csn}
It has been observed that long sequences of contour segments typically have a good amount of intersection with the important object boundaries. Therefore, in our approach, we try to extract long sequences of segments or chains. Furthermore, these segments must be connected to each other strongly, i.e. their connecting end points should be close to each other for them to be considered in the same chain. To this end, we extract a set of salient contours for each image that have been experimentally found to be good candidates for object boundaries and then propose a technique to group these contours into meaningful sequences that have a good chance of overlapping with the boundaries of the objects present in an image. 

%~~~~~~~~~~~~~~~~~~~~~~~~~~~~~~~~~~~~~~~~~~~~~~~~~~~~~~~~~~~~~~~~~~~~~~~~~~~~~~~Subsection :Preprocessing ~~~~~~~~~~~~~~~~~~~~~~~~~~~~~~~~~~~~~~~~~~~~~~~~~~~~~~~~~~~~~~~~
\subsubsection{Obtaining Salient Contours}
\begin{figure*}[t]
\centering
\includegraphics[width=1\linewidth]{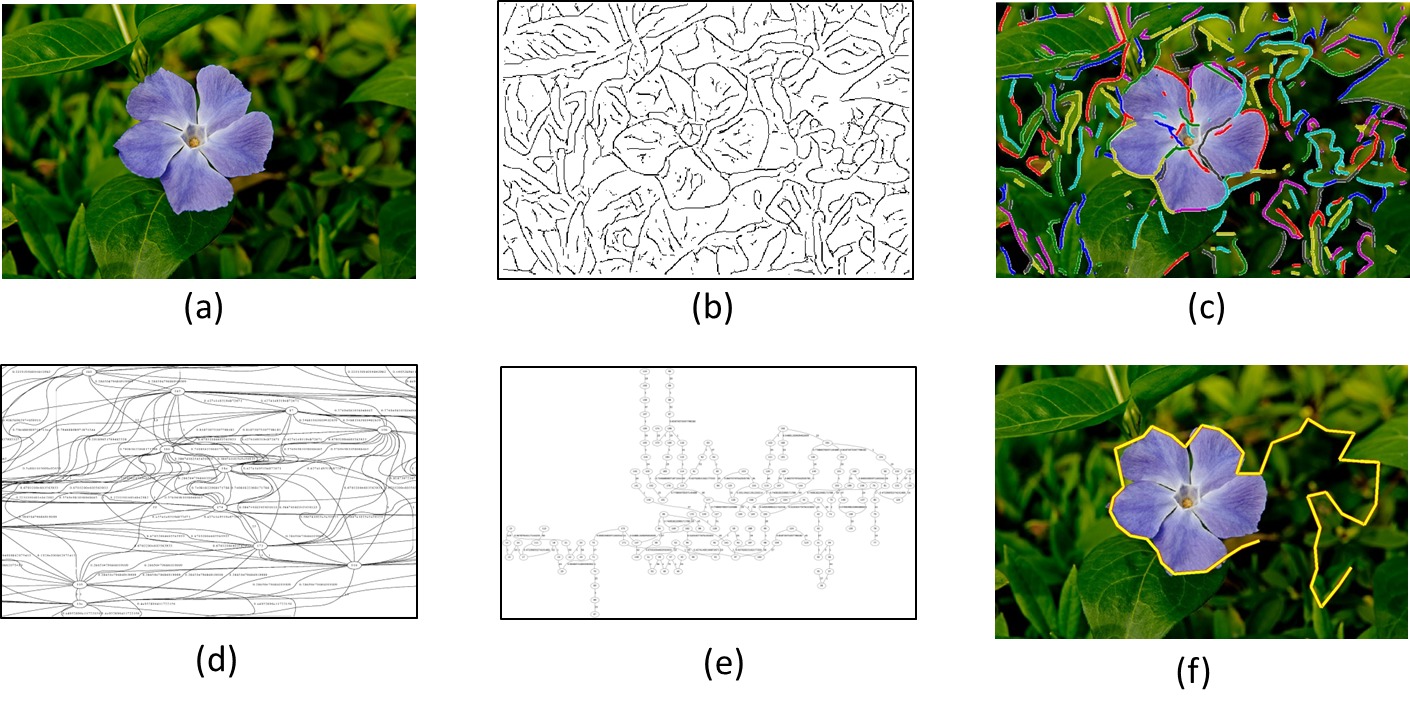} 
%\vspace{-7pt}
\caption{Grouping Salient Contours into Chains: (a) Image, (b) Edgemap~\citep{martin2004learning}, (c) Salient contours~\citep{zhu2007untangling}, (d) Illustrative snapshot of a constructed graph, (e) Maximum Spanning Tree for one component, (f) A long chain}
%\vspace{-10pt}
\label{fig:edgemap_ccut_graph_mst}
\end{figure*}

At first, in order to use fixed parameters, all the database images are normalized to a standard size (we consider the size of the longest side as 256 pixels). Then, the Berkeley Edge Detector~\citep{martin2004learning} is used to generate a probabilistic edge-map corresponding to the object boundaries in the image. This gives superior object boundaries compared to traditional edge detection approaches such as Canny~\citep{canny1986computational} by considering texture along with color and brightness in an image and specifically learning for object boundaries. However such a boundary-map typically still contains a lot of clutter (Fig.~\ref{fig:edgemap_ccut_graph_mst}(b)). Therefore, an intelligent grouping of edge pixels is done to yield better contours that have a higher chance of belonging to an object boundary. The method proposed by~\cite{zhu2007untangling} groups edge pixels by considering long connected edge sequences that have as little bends as possible, especially at the junction 
points as it has been found that the object boundaries mostly follow a straight path at the junction points. Contours that satisfy such a constraint are called \emph{salient} contours in their work and this method is used by us to extract a set of \emph{salient} contours from a database image (Fig.~\ref{fig:edgemap_ccut_graph_mst}(c)).

%~~~~~~~~~~~~~~~~~~~~~~~~~~~~~~~~~~~~~~~~~~~~~~~~~~~~~~~~~~~~~~~~~~~~~~~~~~~~~~~End of Subsection :Preprocessing ~~~~~~~~~~~~~~~~~~~~~~~~~~~~~~~~~~~~~~~~~~~~~~~~~~~~~~~~~~~~~~~~

%-------------------------------------FIG: Edgemap, CCUT, Graph Creation and Spanning Tree---------------------------------
%------------------------------------------------------------------------------------------------

% ~~~~~~~~~~~~~~~~~~~~~~~~~~~~~~~~~~~~~~~~~~~~~~~~~~~~~~~~~~~~~~~~~~~~~~~~~~~~~~~Subsection : Creating segments ~~~~~~~~~~~~~~~~~~~~~~~~~~~~~~~~~~~~~~~~~~~~~~~~~~~~~~~~~~~~~~~~
\subsubsection{Segmenting Salient Contours}

Given the salient contours of an image, we consider bends in them (which remain since the object shape has bends). 
\iffalse The salient contours thus obtained may still contain some bends in them. \fi
Some articulation should be allowed at such bends since it has been observed that the object shape perspective remains relatively unchanged under articulations at such bend points~\citep{basri1998determining}. These bend points along the contour are determined as the local maxima of the curvature. Although curvature has been defined in the literature in many ways~\citep{cohen1992tracking,basri1998determining,chetverikov2003simple}, we use a simple formulation that is fast and robust. The curvature of a point $p_c$ is obtained using $m$ points on either side of it as:
\begin{equation}
\kappa_ {p_c} = \sum_{i=1}^m w_i \cdot  \angle p_{c-i}p_cp_{c+i}
\label{HCPDetection}
\end{equation}
where $w_i$ is the weight defined by a Gaussian function centered at $p_c$. This function robustly estimates the curvature at point $p_c$ at a given scale $m$.
The salient contours thus obtained are split into multiple segments at such high curvature points and as a result, a set of straight line-like segments are obtained for an image.\iffalse (Fig.~\ref{fig:contourChainCreationFlow}(d)).\fi

% ~~~~~~~~~~~~~~~~~~~~~~~~~~~~~~~~~~~~~~~~~~~~~~~~~~~~~~~~~~~~~~~~~~~~~~~~~~~~~~~End of Subsection : Creating segments ~~~~~~~~~~~~~~~~~~~~~~~~~~~~~~~~~~~~~~~~~~~~~~~~~~~~~~~~~~~~~~~~

% ~~~~~~~~~~~~~~~~~~~~~~~~~~~~~~~~~~~~~~~~~~~~~~~~~~~~~~~~~~~~~~~~~~~~~~~~~~~~~~~Subsection : Chaining the segments ~~~~~~~~~~~~~~~~~~~~~~~~~~~~~~~~~~~~~~~~~~~~~~~~~~~~~~~~~~~~~~~~
\subsubsection{Chaining the Segments}
\label{sec:chainingSegments}
\begin{figure*}[t]
\centering
\subfloat[][]{\includegraphics[width=0.35\linewidth]{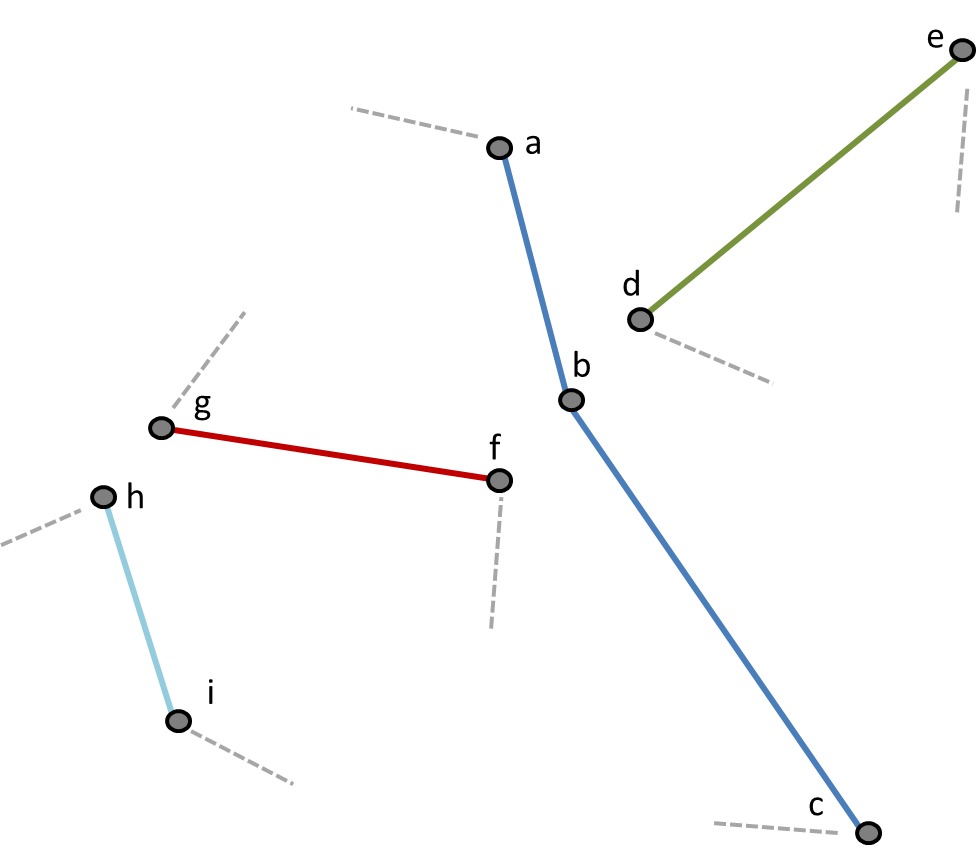}\label{a}}
\qquad
\subfloat[][]{\includegraphics[width=0.35\linewidth]{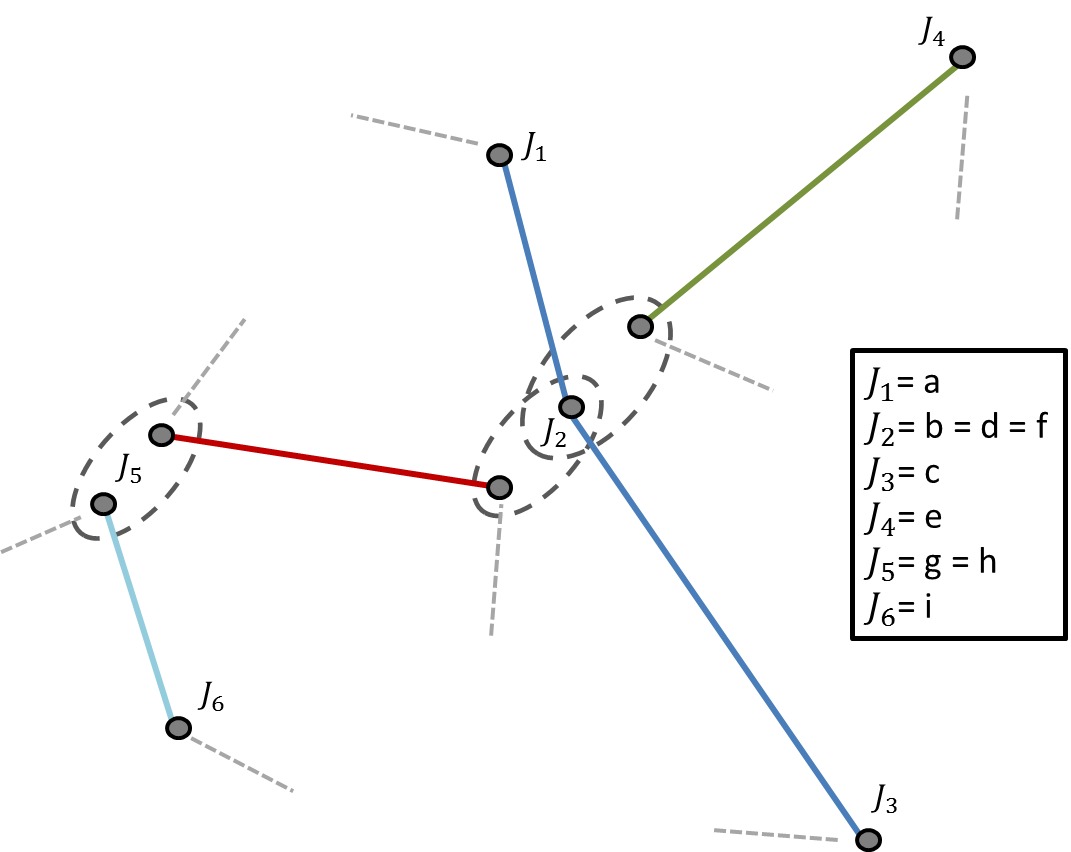}\label{b}} 
%\vspace{-7pt}
\caption{Graph of Joints: (a) Set of segments after breaking the salient contours (shown in different colors) at high curvature points (e.g joint b is a high curvature point). (b) Nearby endpoints of two different segments are merged if they are sufficiently close and a graph of joints ($j$) is created.} % combined by creating new joints (e.g J(b,f) or J(f, g)). x can be any joint in this illustration.}
%\vspace{-10pt}
\label{fig:segmentsAndJointCreation}
\end{figure*}

Given a set of straight line-like segments in an image, we try to connect them. \iffalse into chains that might be able to cover the object boundaries.\fi The connectivity among the segments suggests an underlying graph structure. Thus, a weighted graph/contour segment network is constructed where each end of a contour segment is considered as a vertex/joint and the edge weight between any two vertices is equal to the length of segment. Vertices from two different contour segments are merged if they are spatially close. \iffalse However, since the gaps between such vertices are very small, we do not assign any weight for these connections. \fi Fig.~\ref{fig:segmentsAndJointCreation}(b) shows the graph corresponding to the illustrative set of straight line-like segments in Fig.~\ref{fig:segmentsAndJointCreation}(a). 
%------------------------------------------------------------------------------------------------

\begin{figure*}[t]
\centering
\subfloat[][]{\includegraphics[width=0.35\linewidth]{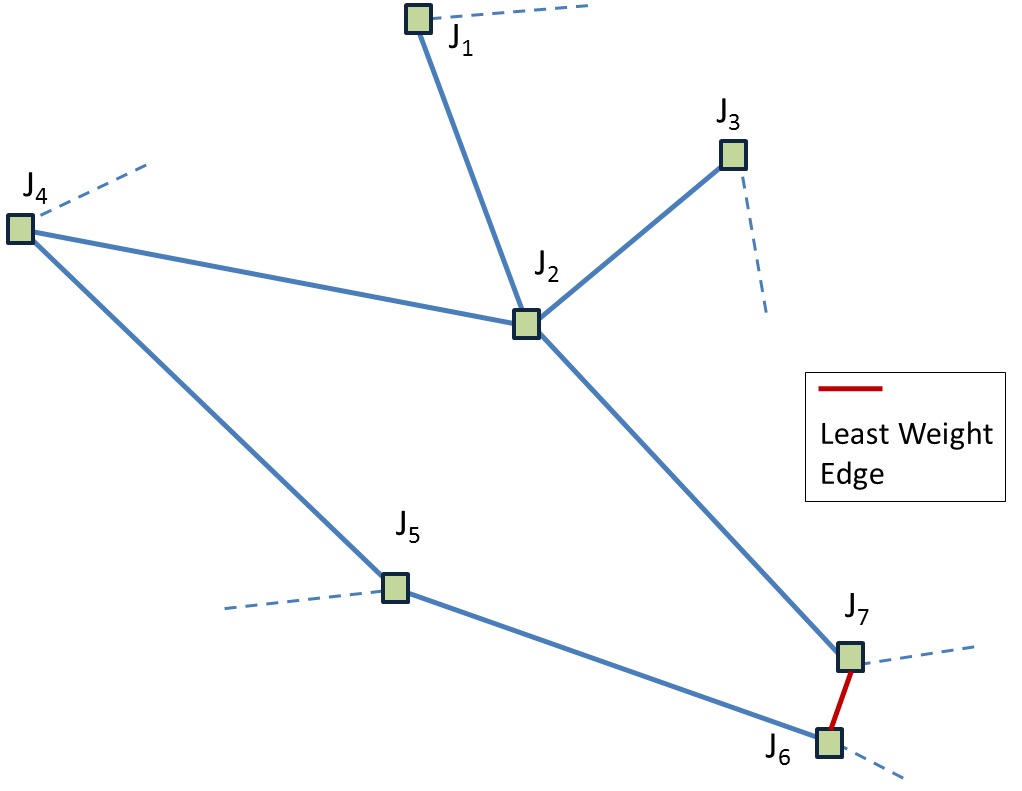}\label{a}}
\qquad
\subfloat[][]{\includegraphics[width=0.35\linewidth]{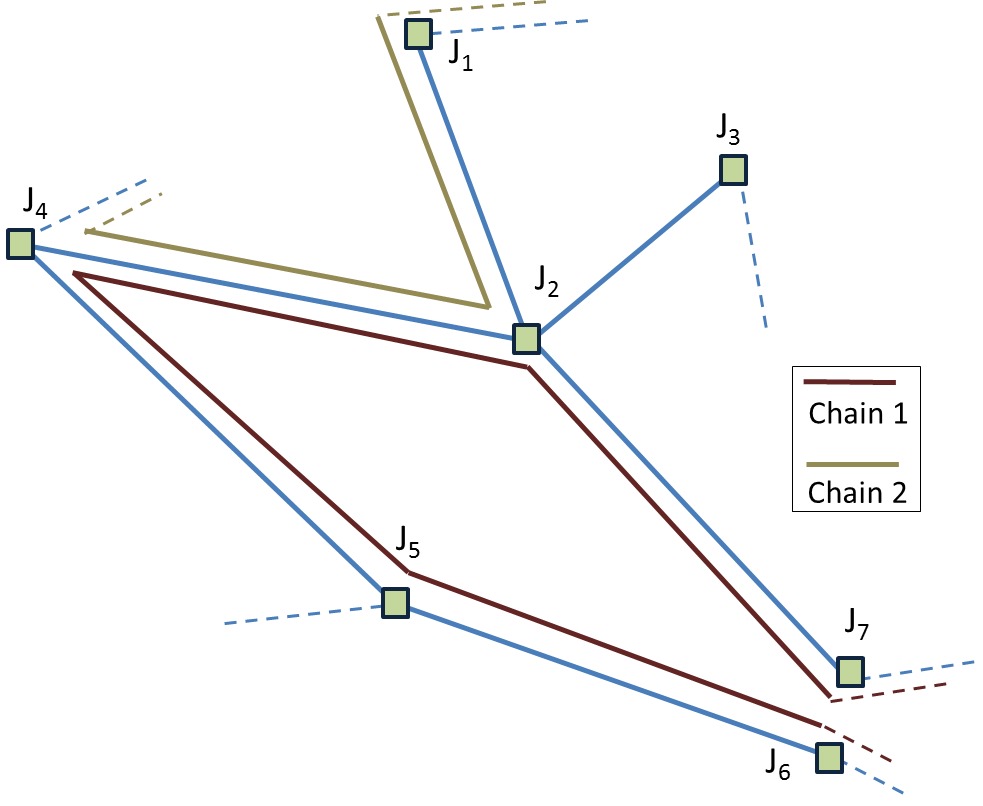}\label{b}} 
%\vspace{-7pt}
\caption{(a) An illustrative snapshot of a network of joints. The red colored edge joining $J_6$ and $J_7$, being the least weight edge in the induced cycle, is removed in the Maximum Spanning Tree of the network. (b) Two possible long paths/chains in the constructed tree.}
%\vspace{-10pt}
\label{fig:jointsGraphAndLongPaths}
\end{figure*}

The weight of an edge in the graph/contour segment network represents the spatial extent of the segments. \iffalse and the connectedness between them. \fi Therefore, a long path in the graph based on the edge weights (higher weight is better) relates to a long connected sequence of contour segments or chains in an image. As the graph may contain cycles, to get non-cyclical paths, the maximum spanning tree\footnote{Maximum spanning tree of a graph can be computed by negating the edge weights and computing the minimum spanning tree.} is constructed for each connected component in the contour segment network (Fig.~\ref{fig:edgemap_ccut_graph_mst}(e)) using a standard minimum spanning tree algorithm~\citep{cormen2009introduction} and paths are extracted from these trees. Note that the Maximum Spanning Tree algorithm removes the minimum weight edges in any cycle (Fig.~\ref{fig:jointsGraphAndLongPaths}(a)). 
%-------------------------------------FIG: JointsGraph,LongPaths---------------------------------
%------------------------------------------------------------------------------------------------
Since we consider only long chains, minimum weight edges which correspond to the small segments would not be typically picked anyway in a long chain. For instance, in Fig.~\ref{fig:jointsGraphAndLongPaths}(a), a long chain passing through the joints $J_7$ and $J_6$ follows the path $< ..., J_7, J_4, J_5, J_6, ...>$ and therefore, removal of the smallest segment $\overline{J_7J_6}$ in the cycle does not lead to any change in the extracted long chain. Fig.~\ref{fig:jointsGraphAndLongPaths}(b) illustrates possible chains for the contour segment network in Fig.~\ref{fig:jointsGraphAndLongPaths}(a) while Fig.~\ref{fig:edgemap_ccut_graph_mst}(f) shows an extracted chain for an image (Fig.~\ref{fig:edgemap_ccut_graph_mst}(a)).

\begin{comment}
A long path or chain thus obtained may deviate from the object boundary at the junction points (Fig.~\ref{fig:edgemap_ccut_graph_mst}(f)). Ideally, to capture maximum shape information, all possible sequences through such junction points should be considered. However, this drastically increases the number of chains in the representation and is therefore impractical for a database of millions of images. 
Hence, as a trade-off between the representative power and compactness, a greedy approach may be followed by considering only edge-disjoint long sequences in the graph. These can be determined by sequentially finding and removing the longest paths in the graph. As a result, an image may be represented by a set of strictly non-overlapping chains.
\iffalse This leads to an image representation by a set of strictly non-overlapping chains. \fi
This representation is compact and typically results in distinctive long chains as shown in our prior work~\cite{parui2014similarity}.
\end{comment}

%-------------------------------------FIG: Reasoning Behind Multiple Chain ---------------------------------
\begin{figure*}[t]
\centering
\includegraphics[width=1\linewidth]{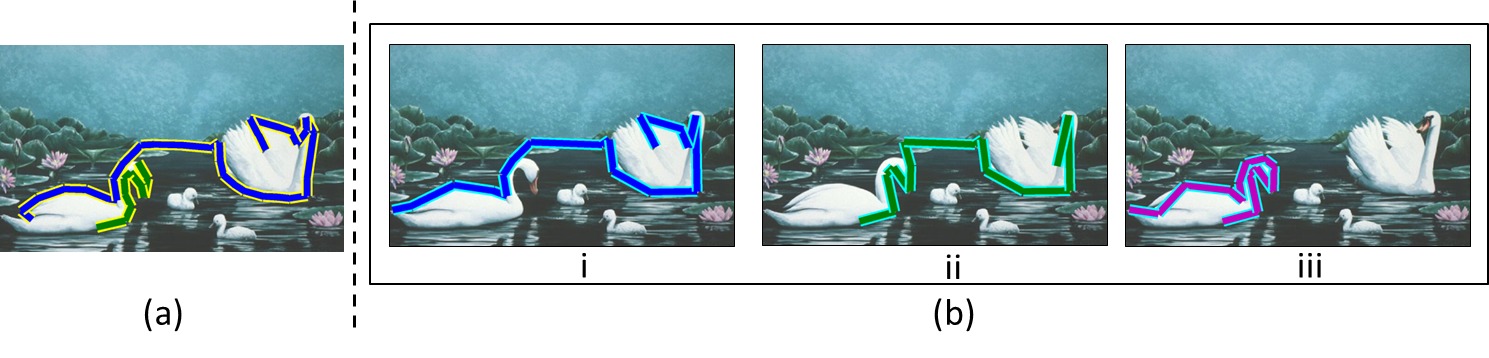} 
%\vspace{-7pt}
\caption{(a) A single chain does not capture the object boundary, which can be addressed by considering multiple overlapping chains \iffalse allowing certain overlap among them\fi (b).}
%\vspace{-10pt}
\label{fig:multipleChainRequirement}
\end{figure*}
%------------------------------------------------------------------------------------------------

\begin{figure*}[t]
\centering
\includegraphics[width=0.15\textwidth]{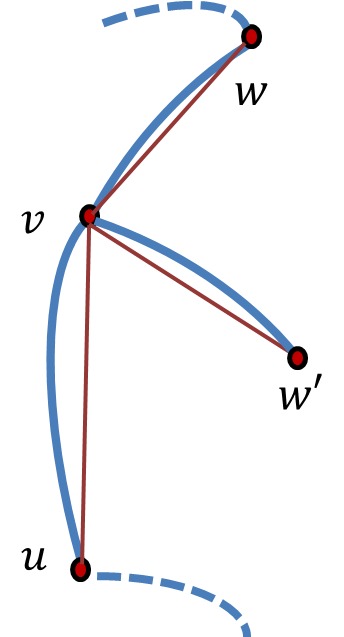}
%\vspace{-7pt}
\caption{$\left\langle ... uvw ... \right\rangle$, being comparatively smoother, is preferred to $\left\langle ... uvw^\prime ...\right\rangle$ for scoring possible paths through the junction $v$}
%\vspace{-10pt}
\label{fig:jointSmoothnessIllustration}
\end{figure*}

A single long chain from this tree typically cannot cover the distinctive portion of any object (Fig.~\ref{fig:multipleChainRequirement}(a)). Therefore, to extract informative chains, the best $N_{OC}$ chains are determined by allowing certain overlaps among the chains. The object boundary is typically smooth. Therefore, in the case of multiple possibilities at the junction points, we measure the smoothness of possible paths at that junction and prefer an almost straight path compared to a curve. Fig.~\ref{fig:jointSmoothnessIllustration} illustrates two possibilities at the junction $v$ and the sequence $\left\langle ... uvw ... \right\rangle$ is preferred to $\left\langle ... uvw^\prime ...\right\rangle$. To this end, the angle between three successive joints $u, v\text{ and } w$ is calculated. Since a straight line is preferred, the deviation of $\angle uvw$ from 180\textdegree\space is determined as a measure of smoothness. 
Furthermore, to use this measure only in the case of multiple possibilities at junction points, smoothness at a joint of a chain is normalized by the smoothness terms of all possible chains at that joint. 
Let $t, u, v, w$ be four consecutive joints along a chain $C$ and $\dis(u, v)$ be the distance between any two joints $u$ and $v$. Then the score of the chain $C$ is determined as:
% \begin{equation}
\begin{dmath}
Score(C) = \sum_{\substack{u, v \in C}} \dis(u, v) \cdot \left[ 1 + 
	  \frac{\lambda_{l}}{2} \cdot \frac{\exp\left(-\lambda_{s}\cdot \left| \pi - \angle tuv \right|\right)}{\sum\limits_{x \mid t,u,x \in C^\prime} \exp\left(-\lambda_{s}\cdot \left| \pi - \angle tux \right|\right)} \\
	  + \frac{\lambda_{l}}{2} \cdot \frac{\exp\left(-\lambda_{s}\cdot \left| \pi - \angle uvw \right|\right)}{\sum\limits_{y \mid u,v,y \in C^{\prime\prime}} \exp\left(-\lambda_{s}\cdot \left| \pi - \angle uvy \right|\right)}   
	  \right]	  
\label{OverlappingChainScore}
\end{dmath}
% \end{equation}
Here $\lambda_{l}$ and $\lambda_{s}$ are two scalar constants. $C^\prime$ and $C^{\prime\prime}$ are possible chains through the joints $t, u$ and $u, v$ respectively. Negative exponential functions are used since only the values close to the desired value (i.e. $\pi$) can be considered a good candidate and anything beyond a certain limit should be given a low score. Note that the smoothness term is weighted by the length of the segment in order to achieve robustness with respect to the number of intermediate joints.
A tree with $N_{l}$ number of leaves has $\binom{N_{l}}{2}$ paths and there is a substantial overlap of joints among many paths.
We exploit this and use a Least Common Ancestor-based implementation~\citep{cormen2009introduction} to efficiently score all $\binom{N_{l}}{2}$ paths and to sequentially select some top $N_{OC} ( = 5)$ chains such that the relative overlap among them is less than $\lambda_{thresh_{chain}} (=60\%)$.
Fig.~\ref{fig:multipleChainRequirement}(b) illustrates the usefulness of considering multiple overlapping chains where only the third chain encapsulates the informative shape information of a swan.
Quite reasonable results may be achieved using such long chains in a contour segment network alone~\citep{parui2014similarity}. We next propose another technique that can often provide complimentary chains in the case this algorithm fails.

%-------------------------------------FIG: Smoothness Illustration ---------------------------------

\subsection{Using Segmented Object Proposals}
\label{sec:gop}
\begin{comment}
Partitioning an image into semantically interpretable regions is an interesting topic in Computer Vision and image segmentation techniques are widely used to obtain possibly connected regions. 
\iffalse plays important role in human perception~\cite{wertheimer1923laws,Barrow1978}. Therefore to obtain object regions, image segmentation techniques are widely used. \fi 
These \iffalse perceptual grouping\fi techniques typically consider different image cues, such as brightness, color, texture over local image patches and then cluster these features using statistical techniques such as mixture models~\cite{yang2008unsupervised,carson2002blobworld}, finding modes~\cite{comaniciu2002mean,vedaldi2008quick}, region-based split-and-merge techniques~\cite{panjwani1995markov,haralick1985image,deng1999color}, global optimization approaches~\cite{zhu1996region,kwatra2003graphcut} or graph partitioning algorithms~\cite{shi2000normalized,felzenszwalb2004efficient}. To combine pixels into perceptually meaningful regions, many superpixel-based grouping techniques have been proposed as well~\cite{moore2008superpixel,veksler2010superpixels,achanta2012slic} which typically produce an over-
segmentation of the image. However, for the purpose of object boundary representation, the unit of interest is a single object region for each object present in an image. 
\end{comment}

\iffalse Therefore, instead of considering the algorithms that exhaustively and uniquely label every pixel or superpixel in an image, we consider segmented object proposals.\fi 

The primary objective of popular segmented object proposal techniques~\citep{carreira2012cpmc,uijlings2013selective,arbelaez2014multiscale} is to provide locations and boundaries of the possible objects in an image. Since we need to extract boundary information for millions of images, we use a very fast method called Geodesic Object Proposals (GOP)~\citep{krahenbuhl2014geodesic} for extracting a set of possible object regions.  This method typically produces many overlapping scored object regions. To limit the number of chains for each image, we consider only some top $N_{GOP} (= 20)$ proposals based on their scores.

\begin{comment} %%%%%%%%%%%%%% TO BE INCLUDED IN THESIS%%%%%%%%%%%%%%%%
 
In GOP, the authors first over-segment the image~\cite{dollar2013structured} into a set of superpixels. 
With an objective of automatically selecting object seeds, a geodesically central superpixel is chosen as the initial seed and then the next seeds are placed far from the existing seeds.
RankSVM~\cite{joachims2002optimizing} is used as a classifier to train the system for better seed placement. For an individual seed, a foreground and background mask is generated in order to compute a geodesic distance transform. Level sets of each of the distance transforms define an object segment. This method typically produces many overlapping object regions. These regions are scored based on the growth of the region and the overlap with other regions. To limit the number of chains for each image, we consider only some top $N_{GOP} (= 20)$ proposals based on their scores.
\end{comment}

\begin{figure*}[t]
\centering
\subfloat[][]{\includegraphics[width=0.5\linewidth]{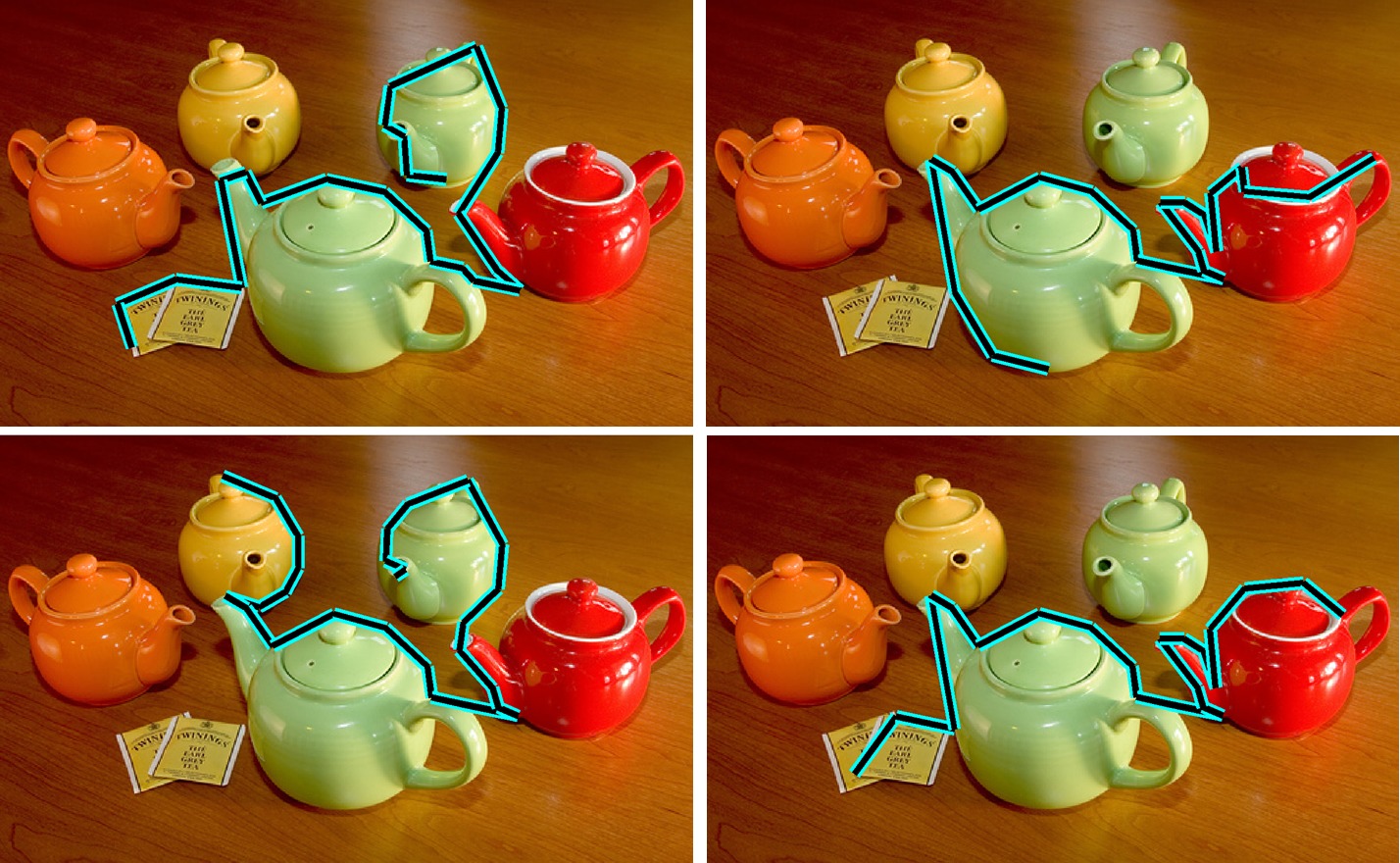}\label{a}}
\ 
\subfloat[][]{\includegraphics[width=0.49\linewidth]{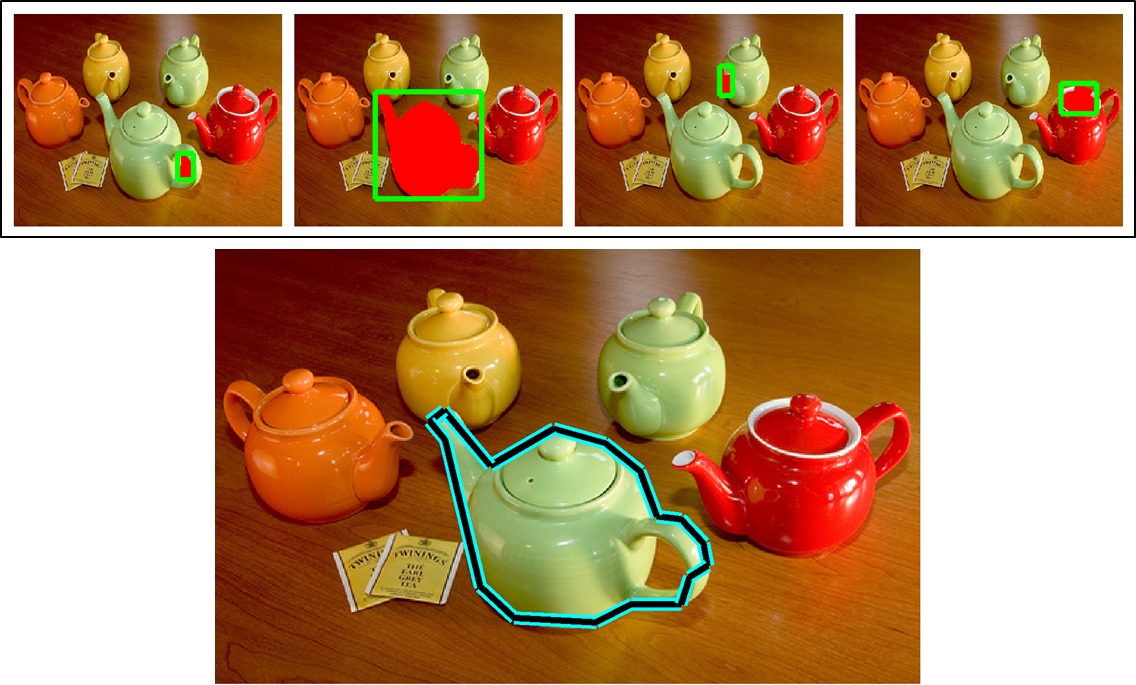}\label{b}} 
%\vspace{-7pt}
\caption{Limitation of considering only top $N_{oc}$ chains from contour segment network: (a) None of the chains encloses sufficient object boundary, (b) A distinctive chain is extracted (below) considering segmented object proposals~\citep{krahenbuhl2014geodesic} (above).}
%\vspace{-10pt}
\label{fig:failureOfMultipleChain}
\end{figure*}

\begin{figure*}
\centering
% \subfloat[][]{\includegraphics[width=0.9\linewidth]{illustrations/GOPFailure/giraffe/GoPFailure_giraffe_banal2_gop}\label{a}}\\
% \subfloat[][]{\includegraphics[width=0.9\linewidth]{illustrations/GOPFailure/giraffe/GoPFailure_giraffe_banal2_MultipleChainFull}\label{b}}\\
\subfloat[Top segmented proposals obtained using the default parameters of~\cite{krahenbuhl2014geodesic} fail to capture the object shape]{\includegraphics[width=1\linewidth,height=3.5	in]{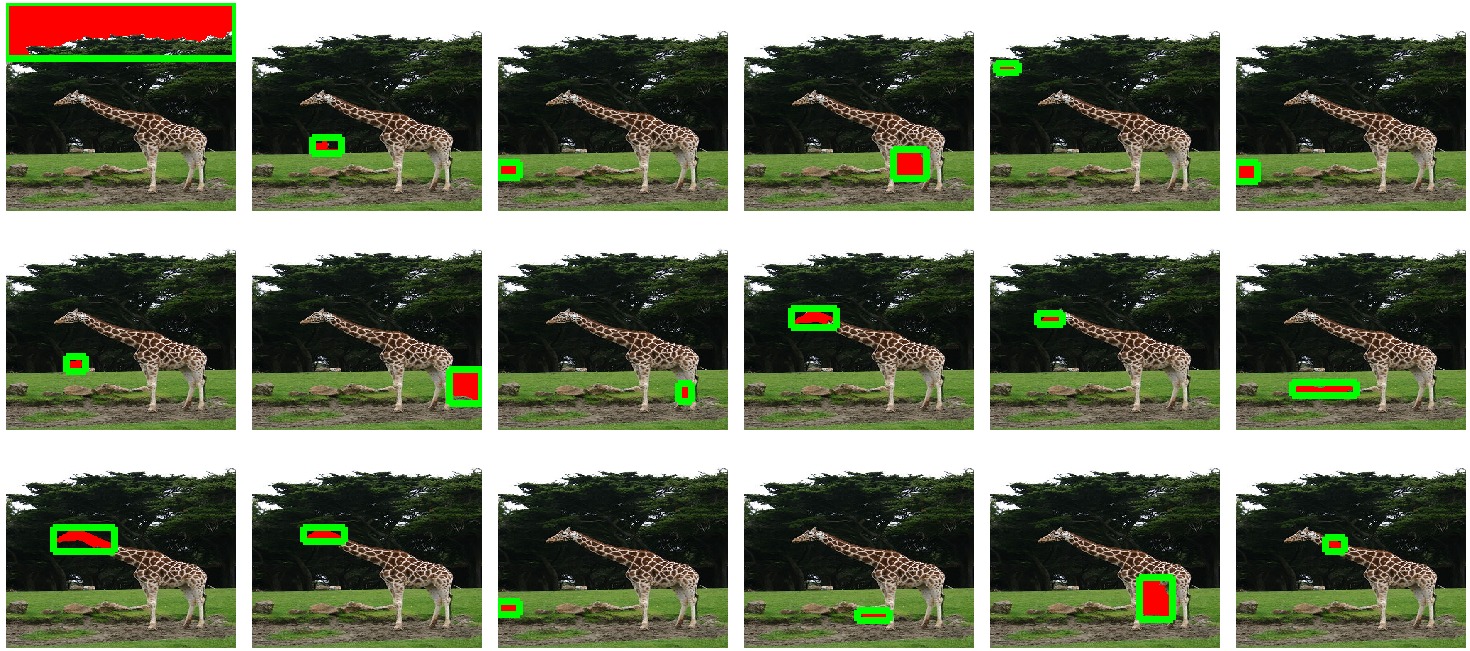}\label{a}}\\
\subfloat[Overlapping chains extracted using contour segment network cover distinctive object boundary]{\includegraphics[width=0.7\linewidth]{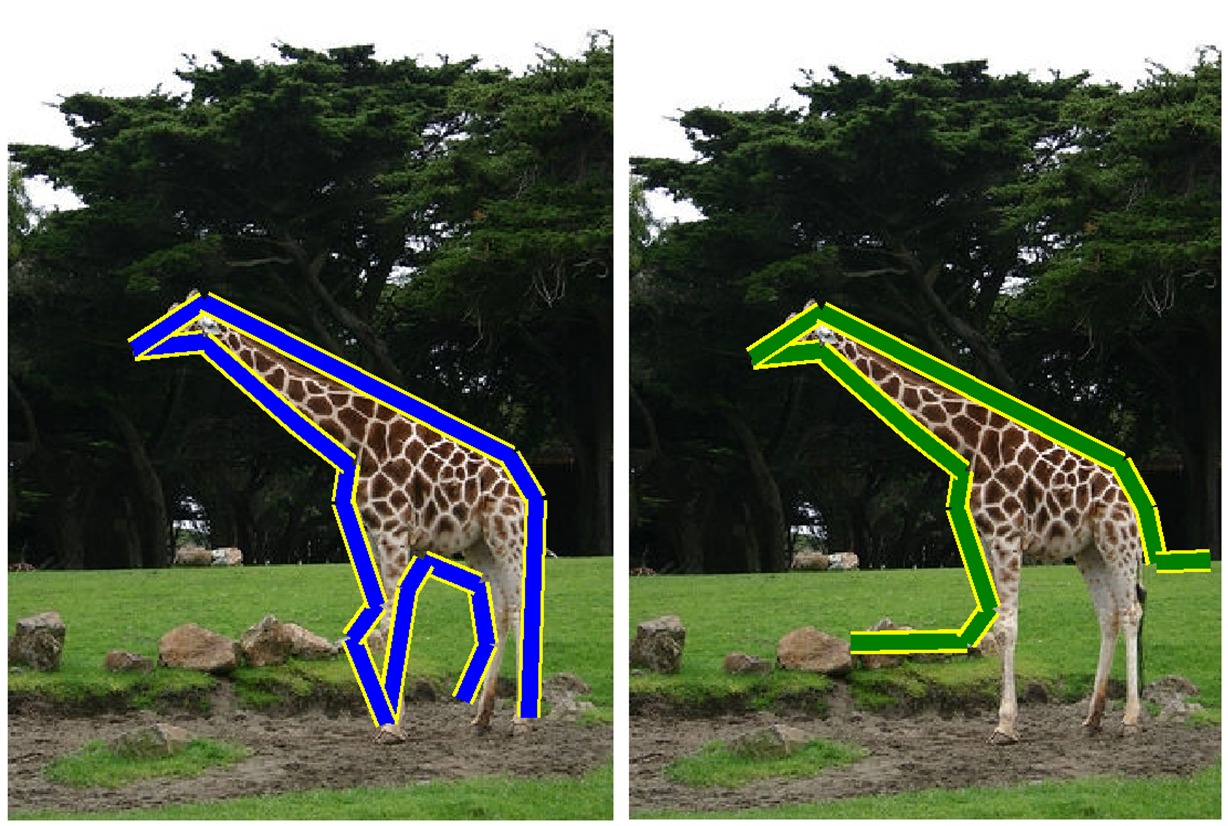}\label{b}}\\
%\vspace{-7pt}
% \caption{Top proposals may fail to capture the object shape (above), whereas chaining the salient contours leads to successful object boundary extraction (below)}
\caption{Limitation of extracting chains only from boundaries of segmented object proposals}
%\vspace{-10pt}
\label{fig:GOPFailure}
\end{figure*}

Note that while it is possible to use image segmentation techniques such as~\cite{shi2000normalized} and~\cite{arbelaez2011contour} to obtain good connected regions, they are typically very slow~\citep{uijlings2013selective}.
\iffalse However, the state-of-the-art segmentation techniques are typically very slow~\citep{uijlings2013selective}. To combine pixels into perceptually meaningful regions, many superpixel-based grouping techniques have been proposed as well~\citep{moore2008superpixel,veksler2010superpixels,achanta2012slic} which typically produce an over-segmentation of the image. However, \fi
Furthermore, for the purpose of object boundary representation, the unit of interest is a single object region for each object present in an image. Therefore, Geodesic Object Proposal~\citep{krahenbuhl2014geodesic} is used in this work. \iffalse that generates a set of object region proposals. \fi
The method is based on superpixel growing and each proposal corresponds to a segment in the image. Thus, the object shape information can be easily extracted by considering the boundaries of the proposed segments. We remove the segments that mostly touch the image boundary as such segments have incomplete boundaries and a possible object is only partially present in the image. Furthermore, to extract only distinctive object boundary, very small regions are also discarded (Fig.~\ref{fig:failureOfMultipleChain}(b) and Fig.~\ref{fig:GOPFailure}(a)). Finally the boundaries of the remaining regions from the top $N_{GOP}$ proposals are taken as chains. 
Fig.~\ref{fig:failureOfMultipleChain}(b) shows a chain successfully obtained from the segmented proposals where the chains extracted using contour segment network were inferior (Fig.~\ref{fig:failureOfMultipleChain}(a)). 

%-------------------------------------FIG: Failure of Multiple Chains ---------------------------------
%------------------------------------------------------------------------------------------------

% However, sometimes a proposed segment contains multiple objects or only a portion of an object. 
Fig.~\ref{fig:GOPFailure} illustrates a reverse example where the top object proposals do not contain informative object boundary information, while the overlapping chains extracted from the contour segment network covers the desired object boundary. Therefore we consider both the approaches for extracting chains for database images in an offline process.

%-------------------------------------FIG: Failure of GOP ---------------------------------
%------------------------------------------------------------------------------------------------

\begin{figure*}[t]
\centering
\includegraphics[width=1\linewidth]{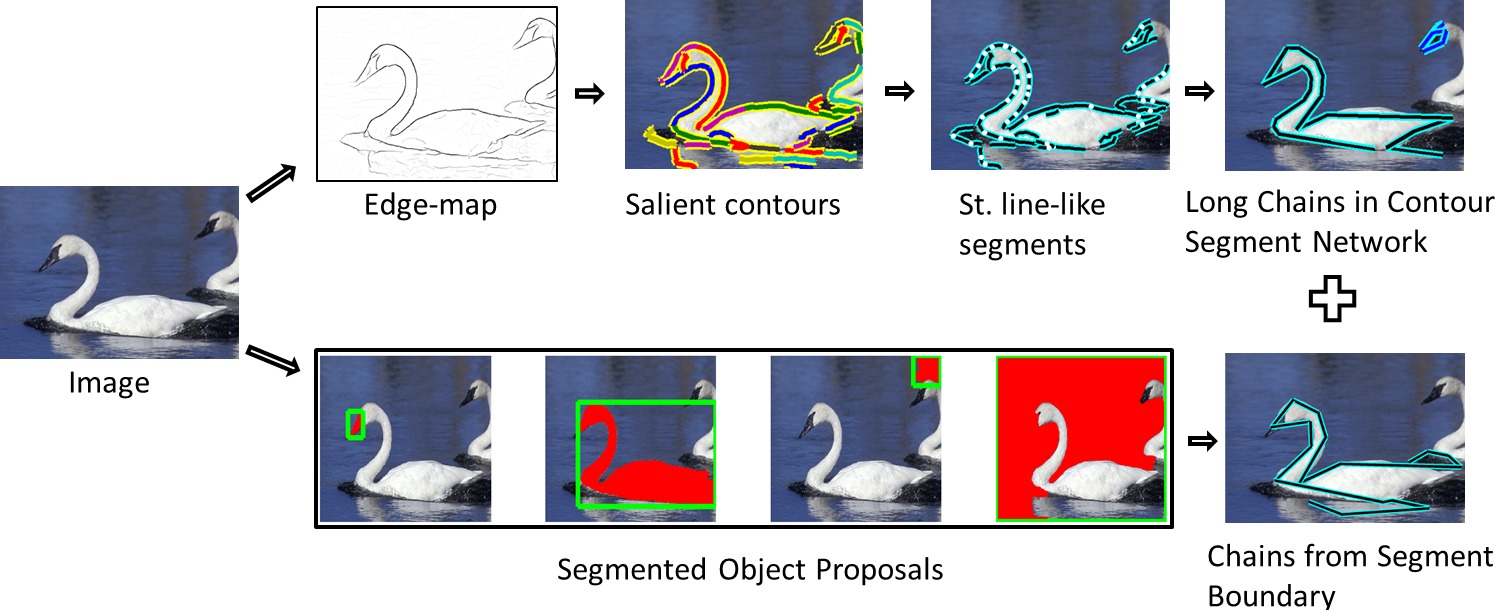} 
%\vspace{-7pt}
\caption{Entire chain creation framework}
%\vspace{-10pt}
\label{fig:CompleteChainCreationFlow}
\end{figure*}

\begin{figure*}	
\centering
\includegraphics[width=0.8\linewidth]{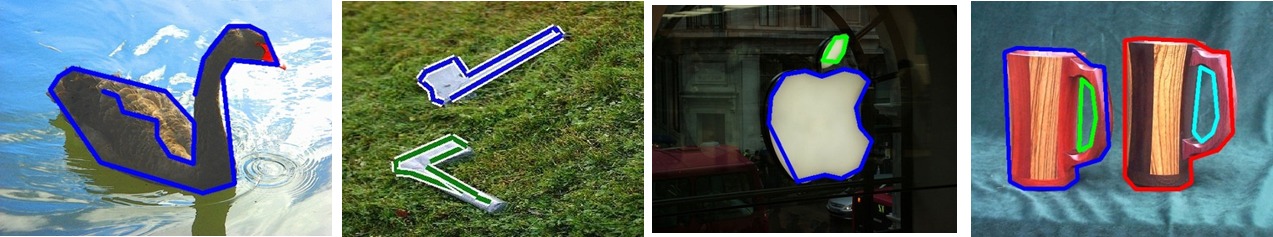}
\caption{Chains extracted for some images. Different chains are represented using different colors. }
\label{fig:contourMaps}
\end{figure*}

Fig.~\ref{fig:CompleteChainCreationFlow} demonstrates the entire chain creation framework and Fig.~\ref{fig:contourMaps} shows the chains thus obtained in some common images. Note that in our framework, it is easy to adapt other possibly better boundary extraction mechanisms as well due to the flexibility of the framework in terms of chain length and the number of chains. 

%-------------------------------------FIG: Chain Creation Flow ---------------------------------

\begin{figure*}	
\centering
\includegraphics[width=0.35\linewidth]{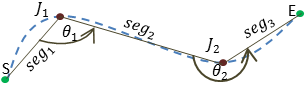} 
\caption{The chain for the curve SE is composed of three line segments. The descriptor for this chain is $\varPsi=\left\langle \gamma_{i}=\frac{l_{seg_{i}}}{l_{seg_{i+1}}}, \theta_i \left|\right. i \in \lbrace1, 2\rbrace \right\rangle$.}
\label{fig:Descriptor}
\end{figure*}

%------------------------------------------------------------------------------------------------------------------
% ~~~~~~~~~~~~~~~~~~~~~~~~~~~~~~~~~~~~~~~~~~~~~~~~~~~~~~~~~~~~~~~~~~~~~~~~~~~~~~~Section : Descriptor ~~~~~~~~~~~~~~~~~~~~~~~~~~~~~~~~~~~~~~~~~~~~~~~~~~~~~~~~~~~~~~~~
\section{Creating Descriptors for each Chain}
\label{sec:descriptor}
In order to efficiently match two chains in a similarity-invariant way, we require a compact descriptor that captures the shape information of the extracted chains in a similarity-invariant way. Towards this goal, the local shape information is captured at the joints in a scale, in-plane rotation and translation invariant way.
For the $i^{th}$ joint of chain $k$ ($J_i^k$), the segment length ratio $\gamma_i=\frac{l_{seg_{i}}}{l_{seg_{i+1}}}$ ($l_{seg_{i}}$ denotes the length of the $i^{th}$ segment) and the anti-clockwise angle $\theta_i$ (range: [0, 2$\pi$]) between the adjacent pair of segments $seg_i$ and $seg_{i+1}$ are determined, as shown in Fig.~\ref{fig:Descriptor}. 
The descriptor $\varPsi^{k}$ for a chain $k$ with $N$ segments is then defined as an ordered sequence of such similarity-invariant quantities:
\begin{equation}
\varPsi^{k} = \left\langle \gamma_i, \theta_i \left|\right. i \in \lbrace 1\ldots N-1\rbrace \right\rangle 
\label{eqnDesc}
\end{equation}
Note that~\cite{riemenschneider2010using} also use joint information by measuring the relative angles among all pairs of sampled points along a contour. However, their representation is not scale invariant which leads to a costly online multi-scale matching phase. In contrast, our proposed descriptor is insensitive to similarities and is suitable for efficiently representing and matching contour chain information in millions of images.

Having extracted chains from images and compactly represented them in a similarity-invariant way, we next describe an approach for efficiently matching two such chains.
% ~~~~~~~~~~~~~~~~~~~~~~~~~~~~~~~~~~~~~~~~~~~~~~~~~~~~~~~~~~~~~~~~~~~~~~~~~~~~~~~End of Section : Descriptor ~~~~~~~~~~~~~~~~~~~~~~~~~~~~~~~~~~~~~~~~~~~~~~~~~~~~~~~~~~~~~~~~

%----------------------------------------------------------------------------------------------END OF SECTION 2 --------------------------------------------------------------------------------------

%---------------------------------------------------------------------------------------------SECTION 3 : MATCHING ----------------------------------------------------------------------

\section{Matching Two Chains}
\label{sec:matching_two_chains}
Standard vectorial type of distance measures are not applicable for matching two chains due to the variability in the lengths of the chains in our case.
\iffalse This constraint makes the task more challenging since most of the fast indexing mechanisms for large scale retrieval exploit a metric structure~\cite{datta2008image,jain2010data}. \fi Further, note that the object boundary is typically captured by only a portion of the chain in the database image (Fig.~\ref{fig:CompleteChainCreationFlow}). Therefore, a partial matching strategy of such chains needs to be devised which can be smoothly integrated with an indexing structure to efficiently determine object shape similarity. 

\begin{figure*}[t]	
\begin{center}
\subfloat[][]{\includegraphics[width=0.45\linewidth]{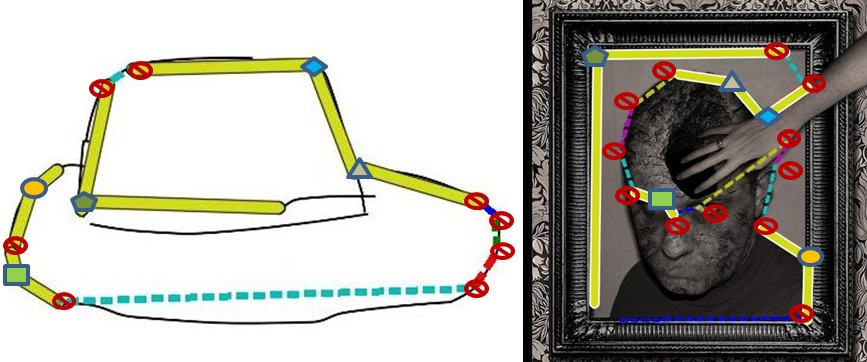}\label{a}}
\qquad
\subfloat[][]{\includegraphics[width=0.45\linewidth]{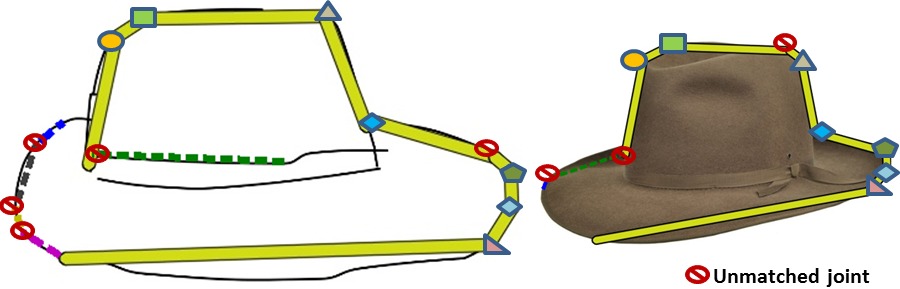} \label{b}}
\end{center}
\caption{(a) A match when fragmented skips are allowed. (b) A match when only almost-contiguous matches are allowed. Matched joints are shown with the same marker in the sketch and the image. Unmatched portions of the chains are indicated by dashed lines.}
\label{fig:intra_chain__hat}
\end{figure*}

Since image chains are typically noisy, it is not uncommon to obtain a chain that captures an object boundary and has non-object contour segments on either side of the object boundary portion. Furthermore, we assume that the object boundary is typically captured by a more or less contiguous portion of the chain without large gaps in between. Although such large split-ups may occur in certain circumstances, allowing such matches leads to a lot of false matches of images due to too much relaxation of the matching criteria. This is illustrated in Fig.~\ref{fig:intra_chain__hat}(a), where the split matches are individually good matches but put together do not match with the intended shape structure at all. Thus, in our work, the similarity between two chains is measured by determining \emph{the maximum (almost) contiguous matching portions of the sequences while leaving out the non-matching portions on either side from consideration} (Fig.~\ref{fig:intra_chain__hat}(b)). 
This is quite similar to the Longest Common Substring\footnote{Substring, unlike subsequence, does not allow gap between successive tokens.}  problem~\citep{cormen2009introduction}, with some modifications that can be solved efficiently using Dynamic Programming. We first consider the individual scores for matching two joints across two chains. 

%-----------------------------------------------------------------FIG: HAT EXAMPLE (FP and TP)------------------------
%INTRA-CHAIN HAT
%-----------------------------------------------------------------------------------------------------------------

%```````````````````````````````````````````````````````````````````SUBSUBSECTION : Joint Similarity ```````````````````````````````````````````````````````````````````
\subsection[short]{Joint Similarity}
 Since exact correspondence of the joints does not capture the deformation that an object may undergo, we provide a slack while matching and score the match between a pair of joints based on the deviation from exact correspondence. The score $S_{jnt}(x,y)$ for matching the $x^{th}$ joint of chain $C_1$ to the $y^{th}$ joint of chain $C_2$ is taken to be the product of two constituent scores:
\begin{equation}
 S_{jnt}(x, y) = S_{lr}(x, y) \cdot S_{ang}(x, y) 
 \label{jointMatchingScore}
\end{equation}
$S_{lr}(x, y)$ is the closeness in the segment length ratio of the two adjacent segments at the $x^{th}$ and $y^{th}$ joints of the two descriptors:
\begin{equation}
%S_{lr}(x, y) = \exp \left( \lambda_{lr} \cdot \left(1 - 1\middle/ \Omega\left(\gamma^{C_1}_{x}, \gamma^{C_2}_{y}\right) \right) \right)
S_{lr}(x, y) = \exp \left( \lambda_{lr} \cdot \left(1 - \middle. \Omega\left(\gamma^{C_1}_{x}, \gamma^{C_2}_{y}\right) \right) \right)
\label{segmentLengthRatio}
\end{equation}
where $\gamma_{x} = \frac{l_{seg_x}}{l_{seg_{x+1}}}$ is as defined in Sec.~\ref{sec:descriptor}, $\Omega\left(a, b\right) = \min\left(a/b, b/a\right), a,b\in\mathbb{R_{\text{\textgreater 0}}}$ measures the relative similarity between two ratios ($\Omega\left(a, b\right) \in (0,1]$) and $\lambda_{lr}(=0.5)$ is a constant. $S_{ang}(x, y)$  determines the closeness of the angles at the $x^{th}$ and $y^{th}$ joints and is defined as:
\begin{equation}
 S_{ang}(x, y) = \exp \left(- \lambda_{ang} \cdot \left|\theta^{C_1}_{x} - \theta^{C_2}_{y}\right|\right)
 \label{jointAngleDifference}
\end{equation}
where $\lambda_{ang}\left(=2\right)$ is a constant.
These two components measure the structure similarity between a pair of joints. 
Due to the consideration of length ratios and relative angles, the joint matching score $S_{jnt}(.,.)$ is also invariant to scale, translation and rotation.
\begin{comment}
However, lengthy segments are more relevant to an object and should get a higher score. Thus, it is desirable to give a higher score to a pair of matched joints if the segment lengths corresponding to the joints are large. 
In general database image chains are noisy, whereas sketch chains are correctly extracted.

However, lengthy segments are more relevant to an object and should get a higher score. Thus, it is desirable to give a higher score to a pair of matched joints if the segment lengths corresponding to the joints are large. This is captured by $S_{sz}$ and is defined as:
\begin{equation}
S_{sz} (x, y) = \min \left( \left(\left. l_{seg_x}^{C_1} + l_{seg_{x+1}}^{C_1}\right)\middle.\right., \left(\left. l_{seg_y}^{C_2}+ l_{seg_{y+1}}^{C_2}\right)\middle. \right. \right)
 \label{weightFactorInJointMatchScore}
\end{equation}
where, $seg_x$ and $seg_{x+1}$ are the two segments on either side of a joint $x$. The information about individual segment lengths is also retained in the chain extraction stage for such a calculation.
\end{comment}
%```````````````````````````````````````````````````````````````````End of SUBSUBSECTION : Joint Similarity ```````````````````````````````````````````````````````````````````

%```````````````````````````````````````````````````````````````````SUBSECTION : Chain Matching ```````````````````````````````````````````````````````````````````
\subsection[short]{Handling Skips}
\label{sec:chain_matching}
Given the scoring mechanism between a pair of joints, the match score between two chains can be determined by calculating the cumulative joint matching score of contiguous portions in the two chains. Although exact matching of such portions can be considered, due to intra-class shape variations, small partial occlusion or noise, a few non-object joints may occur in the object boundary portion of the chain. To handle these non-object portions, some skips need to be allowed. Thus, the problem is formulated as one that finds the longest \emph{almost-contiguous} matching portion of the two chains that are to be matched. Since only descriptors are available at this stage, this matching is performed in the space of chain descriptors.

\begin{figure*}[t]
\centering
\includegraphics[width=0.3\linewidth]{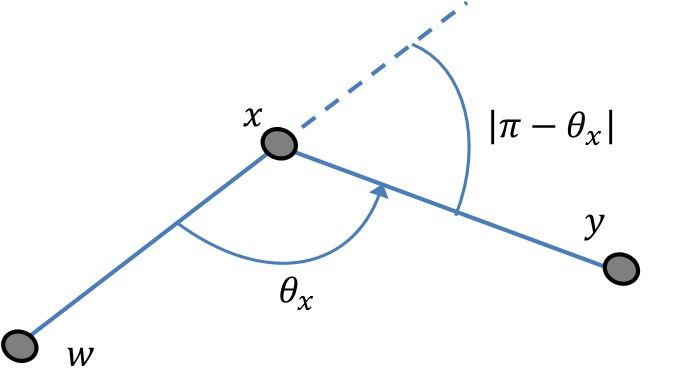} 
%\vspace{-7pt}
\caption{Sharpness ($S_x$) of joint $x$ is calculated by determining the difference of joint angle ($\theta_x$) from $\pi$.}
%\vspace{-10pt}
\label{fig:skipPenaltyAngleIllustration}
\end{figure*}

\iffalse To find the longest match while minimizing the number of skips,\fi
To this end, a skip penalty $\alpha$ is considered for the skipped joints. Note that, the loss of shape information due to a skip depends on the complexity of the skipped joints. It has been observed that a sharper angle captures more shape information than a smoother one. Hence, a skipped joint with a sharper angle should be penalized more.
The sharpness ($S_x$) of any joint $x$ can be calculated by taking the deviation of the joint angle ($\theta_x$) from 180\textdegree\space (Fig.~\ref{fig:skipPenaltyAngleIllustration}):
\begin{equation}
S_x = 1-\exp(-\left| \pi - \theta_x \right|)
\label{eqn:sharpnessOfAngle}
\end{equation}
%-------------------------------------FIG: Skip Penalty Angle Illustration ---------------------------------
%------------------------------------------------------------------------------------------------
Furthermore, lengthier skips typically cause more loss in shape information. Therefore, to penalize skips based on its complexity, along with the sharpness of the skipped angle, the skip penalty ($\omega_x$) is also weighted by the average length of the segments on either side of a skipped joint $x$:
\begin{equation}
\omega_x = S_x + \lambda_{skc} \cdot \frac{\left(l_{seg_x} + l_{seg_{x+1}}\right)}{2}
\label{eqn:skipPenaltyWeight}
\end{equation}
where, to determine the penalty of the skipped segments relative to the chain, the length of each segment is normalized by the length of the chain. $\lambda_{skc}$ is a scalar constant that determines the relative effect of the two components.

\subsection[short]{Matching using Dynamic Programming}
\begin{figure*}[t]
\centering
\includegraphics[width=1\linewidth]{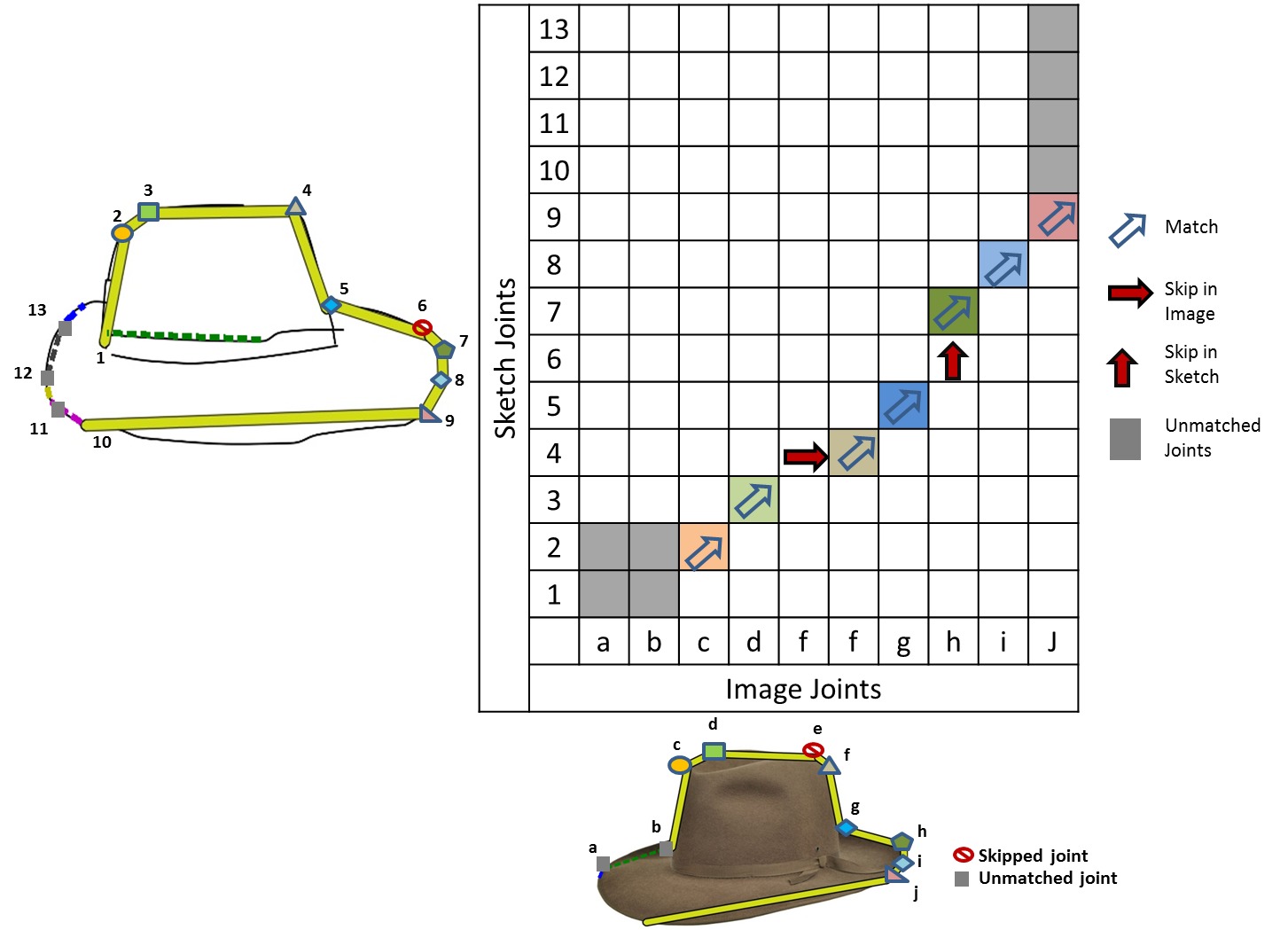} 
%\vspace{-7pt}
\caption{Partial matching of chains with small skips. Matched joints are indicated by the same marker and colored the same in the table (Best viewed in color).}
%\vspace{-10pt}
\label{fig:DP_Illustration}
\end{figure*}

Towards finding almost-contiguous matches, one can formulate the match score $M(p_1,\allowbreak q_1, p_2, q_2)$ for the portion of the chain between joints $p_1$ and $q_1$ in chain $C_1$ and joints $p_2$ and $q_2$ in chain $C_2$. Let the set $\text{J}_1$ and $\text{J}_2$ denote the set of joints of chains $C_1$ and $C_2$ respectively in this interval. Also let $\text{JM}$ be a matching between $\text{J}_1$ and $\text{J}_2$ in this interval. We restrict $\text{JM}$ to obey the order constraint on the matches, i.e., if the joints $a_1$ and $b_1$ of the first chain are matched to the joints $a_2$ and $b_2$ respectively in the second chain, then $a_1$ occurring before $b_1$ implies that $a_2$ also occurs before $b_2$ and vice versa. 
Also let $X(\text{JM})=\left\lbrace x \lvert (x,y) \in \text{JM}\right\rbrace$ and $Y(\text{JM})=\left\lbrace y \lvert (x,y) \in \text{JM}\right\rbrace$ be the set of joints covered by $\text{JM}$.
Then $M(p_1, q_1, p_2, q_2)$ is defined as:
\begin{equation}
\begin{split}
 M(p_1, q_1, p_2, q_2) = \max_{\substack{\text{JM}\in \text{ ordered }\\
		  \text{matchings in} \\
		  \text{the interval}  \\
		  (p_1,q_1)  \text{and} \\
		  (p_2,q_2)}} \left( \sum_{\substack{(x, y) \in \\
								\text{JM}}} S_{jnt}(x, y) - \sum_{\substack{x \in \\
														  \text{J}_1\setminus X(\text{JM})}} \omega^{1}_{x}\alpha^1 - \sum_{\substack{y \in \\
														  \text{J}_2\setminus Y(\text{JM})}} \omega^{2}_{y}\alpha^2 \right) \\    
\end{split}
 \label{matchingScoreWithStartAndEnd}
\end{equation}
Note that $\alpha^1$ and $\alpha^2$ may be different since while matching a sketch chain to an image chain, more penalty is given to a skip in the sketch chain ($\alpha=0.07$) since it is considered cleaner and relatively more free from clutter compared to an image chain ($\alpha=0.03$).
Now, the maximum matching score ending at the joint $q_{1}$ of $C_1$ and $q_{2}$ of $C_2$ from any pair of starting joints, is defined as:
\begin{equation}
 M(q_1, q_2) = \max_{p_1, p_2} M(p_1, q_1, p_2, q_2)
 \label{matchingScoreWithEnd}
\end{equation}
We also take the matching score of a null set ($p_1$\textgreater$q_1$ or $p_2$\textgreater$q_2$) as zero which constrains $M(q_1, q_2)$ to take only non-negative values.
Then, it is not difficult to prove that $M$ can be rewritten using the following recurrence relation:
\begin{equation}
\begin{split}
 M(q_1, q_2)=\begin{cases}
		  0, & \text{if } q_1, q_2 = 0 \\
		  \max \begin{cases}
		        M(q_1-1, q_2-1) + S_{jnt}(q_1, q_2) \\
		        M(q_1-1, q_2) -\omega^{1}_{q_1}  \alpha^{1} \\
		        M(q_1, q_2-1) - \omega^{2}_{q_2} \alpha^{2}\\
		        0
		       \end{cases}, &\text{otherwise}
                  \end{cases}
\end{split}
 \label{matchScore}
\end{equation}
This formulation immediately leads to an efficient Dynamic Programming solution that computes $M$ for all possible values of $q_1$ and $q_2$ starting from the first joints to the last ones. A search for the largest value of $M(q_1, q_2)$ over all possible $q_1$ and $q_2$ will then give us the best almost-contiguous matched portions between two chains $C_1$ and $C_2$ in terms of the highest matching score. Fig.~\ref{fig:DP_Illustration} visually illustrates the Dynamic Programming-based matching procedure, where the chains are partially matched and a few joints are skipped while matching. This approach helps us to efficiently obtain a matching score between a pair of chains. Furthermore, to handle an object flip, we match by flipping one of the chains as well and determine the best matching score as the one that gives the highest score between the two directions. We call the final score between two chains $C_1$ and $C_2$ as the Chain Matching Score $CMS(C_1, C_2)$.

%-------------------------------------FIG: DP Illustration ---------------------------------
%------------------------------------------------------------------------------------------------

%-------------------------------------FIG: Requirement of IntraChain Angular Consistency ---------------------------------
\begin{figure*}[t]
\centering
\includegraphics[width=0.8\linewidth]{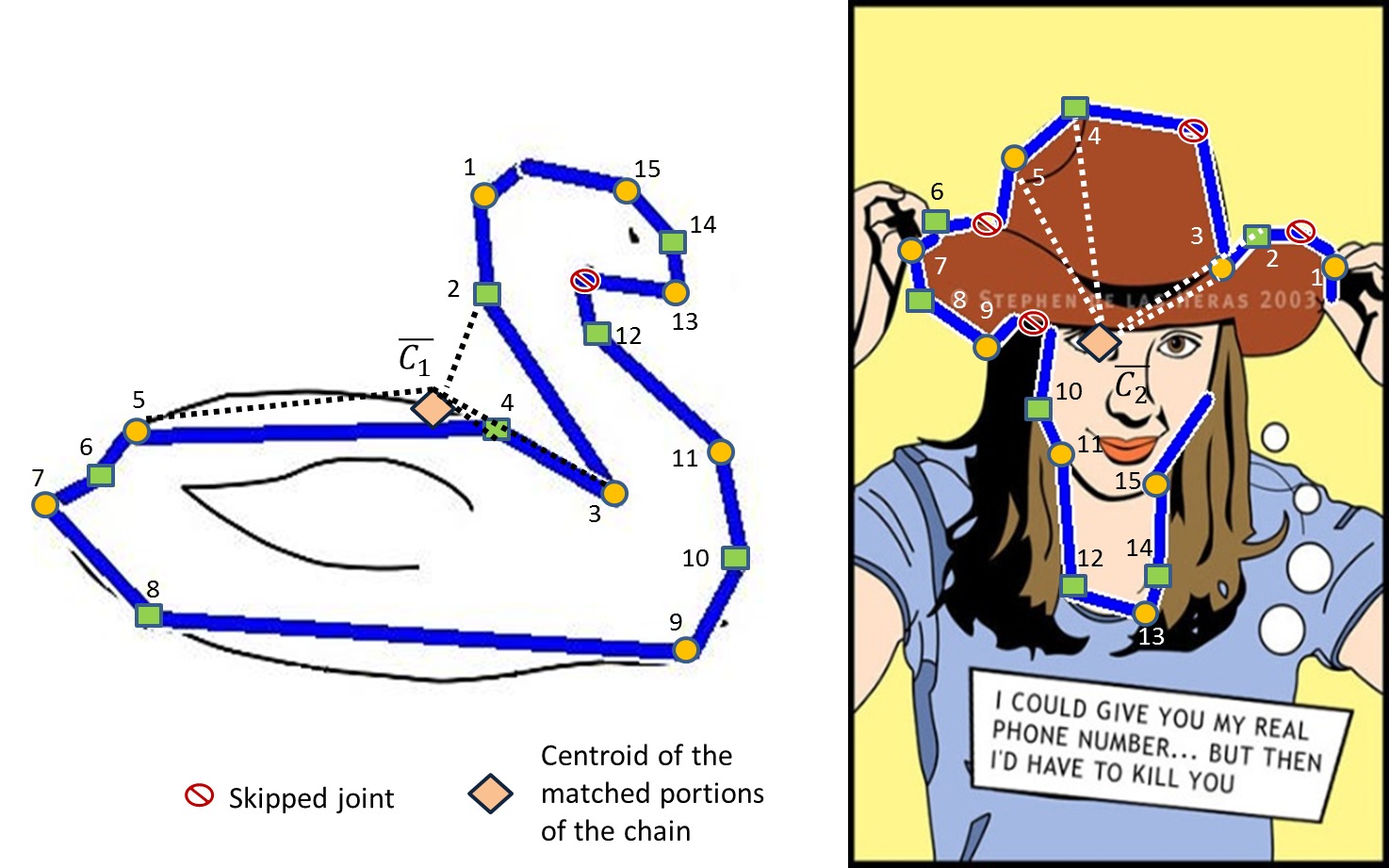} 
%\vspace{-7pt}
\caption{Skipping of important joints can lead to a false positive match. However $\angle2\overline{C_1}3, \angle3\overline{C_1}4, \angle4\overline{C_1}5$ in sketch chain and corresponding angles $\angle2\overline{C_2}3, \angle3\overline{C_2}4, \angle4\overline{C_2}5$ in image chain is highly dissimilar leading to a low Global Angle Consistency score for these two falsely matched chains.}
%\vspace{-10pt}
\label{fig:IntraChainAngularConsistencyReq}
\end{figure*}

The entire operation of matching two chains takes $\mathcal{O}(n_{C_1}*n_{C_2})$ time, where $n_{C_1}$ and $n_{C_2}$ are the number of joints in chains $C_1$ and $C_2$ respectively. It has been observed that a chain typically consists of 12-17 joints leading to a running time of approximately 100-400 units of joint matching, which is not very high. 
Note that, this DP formulation is similar to the Smith-Waterman algorithm (SW)~\citep{smith1981identification}, which aligns two protein sequences based on a fixed alphabet-set and predefined matching costs.~\cite{meltzer2008edge} use SW to perform matching between two images under wide-baseline viewpoint changes. 
Our method is a slight variation from this since it performs matching based on a continuous-space formulation that measures the deviation from exact correspondence to handle deformation.

However, matching two chains by determining local joint correspondence alone sometimes leads to a globally inconsistent match as both deformation and skips of individual joints are allowed while matching. In Fig.~\ref{fig:IntraChainAngularConsistencyReq}, most of the joints are locally matched correctly in a similarity invariant way; but a few joints are skipped in the sketch chain and in the database chain, which leads to a globally inconsistent matching. This necessitates consideration of global consistency of the matched portions of chains for improved matching.

\begin{comment}
However, matching two chains by determining local joint correspondence may lead to a match of globally different shape. This necessitates us to consider global consistency of the matched portions of a pair of chains.

\subsection{Intra-chain Angular consistency}

Individual joints are matched in a scale, translation and rotation invariant way. Furthermore, while matching skipping of joints are allowed in both sketch and in the image. Therefore if some distinctive portion of the object shape gets skipped, it may lead to a false matching of globally dissimilar shape. 
In Fig. X, most of the joints are locally matched in a similarity invariant way. However, a,b,cth joint in database chain and a,b,c th joint in the sketch chain are skipped, which leads to a globally different matching. Thus to maintain global shape similarity, it is necessary to consider the consistency of all the matched joints together. 
\end{comment}

\subsection[short]{Global Angle Consistency of the Matched Chains}
The angle that any two consecutive matched joints make with respect to some global reference point will be similar for two correctly matched chains and different for falsely-matched chains.
% Since this angle difference will be high for globally dissimilar shapes, this angle-consistency measure can be used to refine the matching score between two chains.
\iffalse However, the difference in angle will be high for most of the joints, if two shapes are globally dissimilar. \fi
The centroid of the matched portion of a chain is a robust point that can be used as a reference. Thus, to determine the Global Angle Consistency (GAC) between any two matched chains $C_1$ and $C_2$, we consider the centroid of the matched portion of the chain ($\overline{C}$) as the reference point and calculate the differences in angles that any two consecutive joints make:
\begin{equation}
GAC(C_1, C_2) = \exp\left(-\lambda_{ac}\cdot\frac{1}{N^{J}}\sum\limits_{i=1}^{N^{J}-1} \angle J_{1}^{i}\overline{C_{1}}J_{1}^{i+1} - \angle J_{2}^{i}\overline{C_{2}}J_{2}^{i+1}\right) 
\label{eqn:intraChainGeometricConsistency}
\end{equation}
where, $N^J$ is the total number of matched joints between $C_1$ and $C_2$ and $\lambda_{ac}$ is a scalar constant. A higher value of $\lambda_{ac}$ indicates a harder constraint on the global shape similarity. 
Fig.~\ref{fig:IntraChainAngularConsistencyReq} shows an example where a false match is rejected due to a low global angle consistency score. Note that the computation of the global angle consistency is quite fast as it involves only the matched joints and can be done from the descriptor directly without referring back to the corresponding images.

The chain matching score (CMS) is weighted by the global angle consistency score (GAC) to obtain the final chain similarity score of a pair of chains $C_1 \text{and } C_2$:
\begin{equation}
CS(C_1,C_2) = GAC(C_1, C_2) \cdot CMS(C_1, C_2)
\label{eqn:finalChainScore}
\end{equation}

This chain-to-chain matching strategy is used to match two image chains during indexing as well as a sketch chain to an image chain during image retrieval.
Offline Image Indexing for faster retrieval is considered next.

\section{Image Indexing}
\label{sec:image_indexing_and_fast_retrieval}
Given a chain descriptor, matching it online with all chains obtained from millions of images will take a considerable amount of time. Therefore, for fast retrieval of images from a large dataset, an indexing mechanism is required.
Different indexing techniques have been considered in the literature for content-based image retrieval, viz. tree-based approaches using $k$d tree and its variants~\citep{friedman1977algorithm,vlachos2005rotation,muja2009flann}, hierarchical k-means~\citep{nister2006scalable}, hashing~\citep{indyk1998approximate,deng2011hierarchical,jegou2011product} etc. These approaches exploit the vectorial representation of the extracted features and perform either exact or approximate nearest neighbor search. However, in our representation, the length of each chain is not fixed. Furthermore the matching score cannot be obtained as a direct accumulation of the scores of individual dimensions. It is also not possible to use metric-based indexing techniques in our case due to a violation of the triangle inequality~\citep{keogh2005exact}. These considerations rule out most of the possibilities such as $k$d tree, hashing etc. Therefore, in this work, an approach similar to hierarchical k-means~\citep{muja2009flann,
nister2006scalable} but using medoids instead of means is used, which has been found to perform comparable to the state-of-the-art indexing techniques~\citep{muja2009flann}.

\iffalse Due to the variability in the length of the descriptors, it is difficult to use metric-based data structures, such as \emph{k-d tree}~\cite{muja2009flann} or Vantage-Point tree~\cite{vlachos2005rotation}. Therefore, in this work, an approach similar to \emph{hierarchical k-means}~\cite{muja2009flann,nister2006scalable} is used, \fi

%--------------------------------------------------------------FIG : Index Tree and Requirment of Geometric Consistency------------------------------------------
\begin{comment}
\begin{figure*}[t]	
\begin{center}
\subfloat[][]{\includegraphics[width=0.47\linewidth]{illustrations/HierarchicalIndexingTree/HierarchicalIndexingTree}\label{a}}\qquad
\subfloat[][]{\includegraphics[width=0.47\linewidth]{illustrations/ReqOfGeometricConsistency/ReqOfGeometricConsistency} \label{b}}
\end{center}
\caption{(a) Similar chains are clustered at the leaf nodes of the hierarchical k-medoid-based indexing structure. (b) Considering individual chain matching without global consistency check can lead to a false positive retrieval.}
\label{fig:indexTreeAndReqOfGeometricConsistency}
\end{figure*}
\end{comment}
%---------------------------------------------------------------------------------------------------------------------------

\begin{figure*}[t]
\centering
\includegraphics[width=0.75\linewidth]{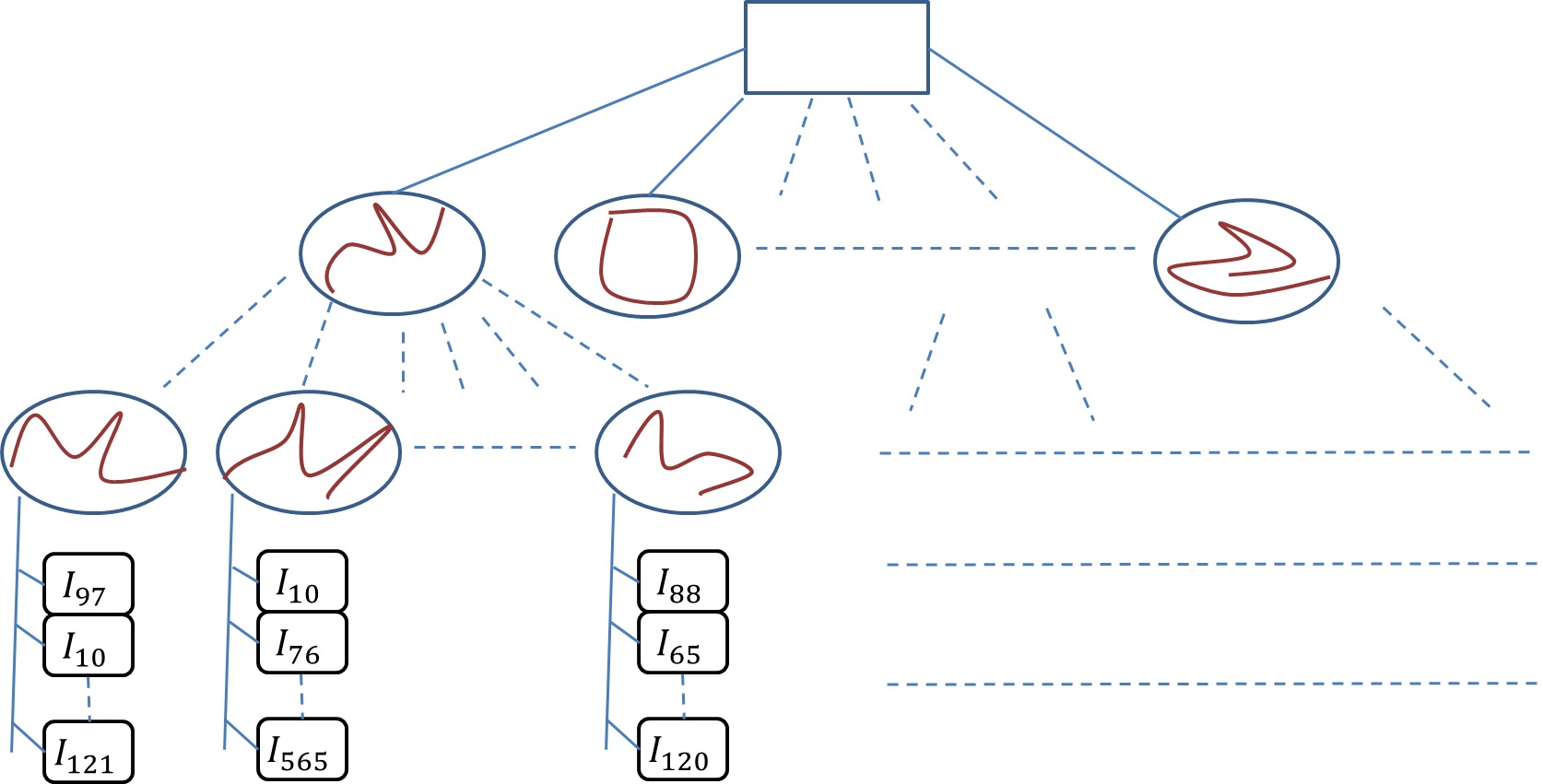} 
%\vspace{-7pt}
\caption{Similar chains are clustered at the leaf nodes of the hierarchical k-medoid-based indexing structure.}
%\vspace{-10pt}
\label{fig:indexTree}
\end{figure*}

All the database chains are considered for indexing and a hierarchical structure is constructed by splitting the set of extracted chains into $k$ different clusters using the \emph{k-medoids} algorithm~\citep{toyama2002probabilistic,opelt2008learning}. \iffalse Note that, because of the variable-length chain descriptors, \emph{k-means} is inapplicable. \fi At first, $k$ chains are chosen as the cluster centroids probabilistically using the initialization mechanism of \emph{k-means++}~\citep{arthur2007k} which increases both speed and accuracy. 
The remaining chains are matched to each medoid chain using Dynamic Programming-based partial matching algorithm and assigned to the closest one based on the matching score (Eqn.~\ref{eqn:finalChainScore}). However, due to partial matching, it is possible to get a high matching score for more than one medoids. Therefore, a chain is assigned to all the medoids for which the matching score is greater than some $Th_{ms}=(80\%)$ of the score of the closest medoid. This operation is then recursively performed on the individual clusters to determine the clusters at different levels of the tree. A leaf node of such a tree maintains a list of images of which at least one chain matches to the corresponding medoid chain. 
Note that, since an image has multiple chains and even one chain can belong to multiple nodes in our approach, the same image can be present at multiple leaves (Fig.~\ref{fig:indexTree}).
Given such a hierarchical chain tree constructed offline during indexing, we next discuss how to search in the Image Database given a query sketch.

%-------------------------------------FIG: Index Tree ---------------------------------
%------------------------------------------------------------------------------------------------
%~~~~~~~~~~~~~~~~~~~~~~~~~~~~~~~~~~~~~~~~~~~~~~~~~~~~~~~~~~~~~~~~~~~~~~~~~~~~~~~~~~~End of Section: Indexing~~~~~~~~~~~~~~~~~~~~~~~~~~~~~~~~~~~~~~~~~~~~~~~~

%~~~~~~~~~~~~~~~~~~~~~~~~~~~~~~~~~~~~~~~~~~~~~~~~~~~~~~~~~~~~~~~~~~~~~~~~~~~~~~~~~~~Begin Section 5: Retrieval~~~~~~~~~~~~~~~~~~~~~~~~~~~~~~~~~~~~~~~~~~~~~~~~
\section{Image Retrieval given a Query Sketch} 
\label{sec:image_retrieval}
 A user typically draws an object along its boundary~\citep{cole2008people}. From a touch-based device, the input order of the contour points of the object boundary is usually available. Therefore, sketch chains can be trivially obtained in an online retrieval system breaking them at turns in the drawing.
 Offline line drawings can be decomposed in a manner similar to the edge-detected Images (Sec.~\ref{sec:chainingSegments}).
 However, chains with less than some $Th_{nj} (= 5)$ joints are discarded as they are very simple and can match to any non-informative portion of another chain. Finally, descriptors are determined in a manner similar to image chains (Eqn.~\ref{eqnDesc}).
%  Finally, descriptors are determined similarly for each of the extracted sketch chains (Eqn.~\ref{eqnDesc}).
 
\subsection{Search in the Hierarchical Chain Tree}  
For each of these sketch chain descriptors, a search is performed in the hierarchical k-medoids tree. At every level of the tree, the query chain is matched with all the medoid chains and then the subtree of the best matched medoid is explored in a best-bin-first manner~\citep{muja2009flann}. At first, a single traversal is performed through the tree following the best matched medoids at every level. This yields a small set of images corresponding to the best matched leaf medoid. 
Since at every level the query chain can get a good match with more than one medoids, to consider those possible matches, all the unexplored branches along the path are added to a priority queue. After the first traversal, the branch closest to the query chain is extracted from this priority queue and explored further. The search procedure stops once a pre-determined number of database images are retrieved. For all these retrieved images, at least one chain of each image matches with the query chain. Note that, for multiple sketch chains, we get multiple sets of images from the leaf nodes of the search tree, all of which are taken for the next step. 

\iffalse  For each of these sketch chain descriptors, a search in the hierarchical k-medoids tree yields a small set of images in which at least one chain for each image matches with the query chain in the tree (Fig. X). Note that, for multiple sketch chains, we get multiple sets of images from the leaf nodes of the search tree, all of which are taken for the next step. \fi

\begin{figure*}[t]
\centering
\includegraphics[width=0.75\linewidth]{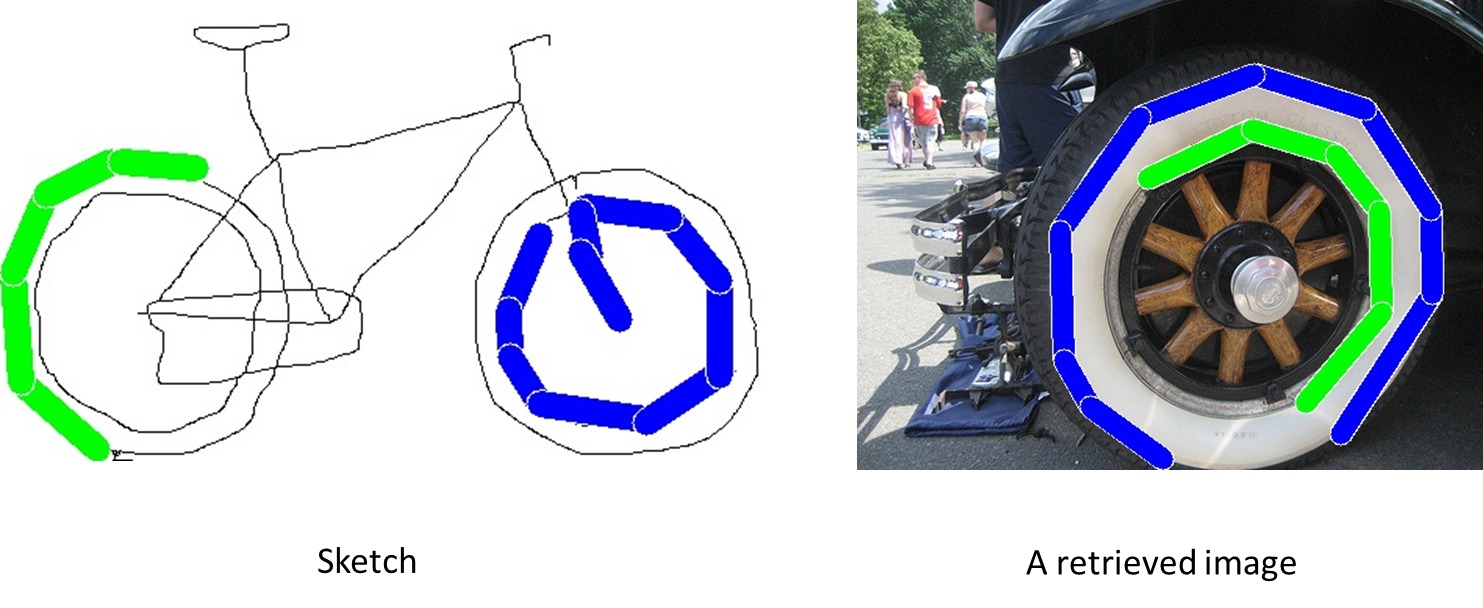} 
%\vspace{-7pt}
\caption{Considering only individual chain matches without global consistency check can lead to a false positive retrieval.}
%\vspace{-10pt}
\label{fig:ReqOfGeometricConsistency}
\end{figure*}

 Given a set of retrieved images with corresponding matched chains, we devise a sketch-to-image matching strategy to rank the images. Since all chain matchings between a sketch and an image may not be retrieved from the hierarchical tree due to low similarity scores, we try to match the remaining chains of the sketch also with other chains of a shortlisted image to obtain the complete chain-matching information between the corresponding sketch and image. 
 The matching score of an image for a given sketch is then calculated based on the cumulative matching scores of individual matched chain pairs between the sketch and the image.
 However, the actual object boundary may be split across multiple chains. Therefore it is necessary to consider all such matchings while determining the match score between a sketch and an image. However, such multiple matches may not be geometrically consistent with each other. Fig.~\ref{fig:ReqOfGeometricConsistency} shows a case where two chains individually match well in both the sketch and the image, but the matches are not geometrically consistent with each other. This necessitates us to consider the geometric consistency of the matched chains to discard false positive retrievals.
 \iffalse Since the actual object boundary may be split across multiple chains, it is necessary to consider geometric consistency of the matched portions of multiple chains for correct retrievals. \fi
 Although such geometric consistency has been studied previously in the literature~\citep{philbin2007object,sattler2009scramsac,tsai2010fast}, it is considered in a new context in this work.
 
   %-------------------------------------FIG: Requirment of Geometric Consistency ---------------------------------
%------------------------------------------------------------------------------------------------

 %~~~~~~~~~~~~~~~~~~~~~~~~~~~~~~~~~~~~~~~~~~~~~~~~~~~~~~~~~~~~~~~~~~~~~~~~~~~~~~~~~~~End of subSection: Retrieval~~~~~~~~~~~~~~~~~~~~~~~~~~~~~~~~~~~~~~~~~~~~~~~~

 %~~~~~~~~~~~~~~~~~~~~~~~~~~~~~~~~~~~~~~~~~~~~~~~~~~~~~~~~~~~~~~~~~~~~~~~~~~~~~~~~~~~Subsection: Geometric Consistency~~~~~~~~~~~~~~~~~~~~~~~~~~~~~~~~~~~~~~~~~~~~~~~~~~~~~~~~~~~
\subsection{Geometric consistency across multiple matched chains}
\label{sec:geometric_consistency_between_matched_chains}
\begin{figure*}[t]	
\begin{center}
\subfloat[][]{\includegraphics[width=0.47\linewidth]{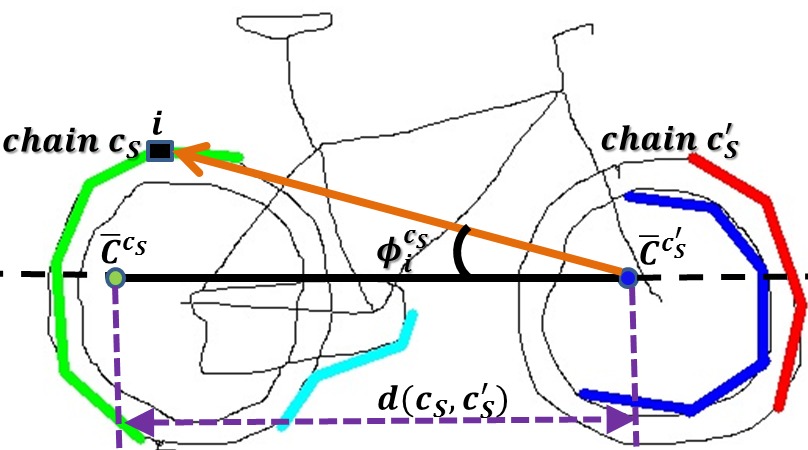}\label{a}}
\subfloat[][]{\includegraphics[width=0.47\linewidth]{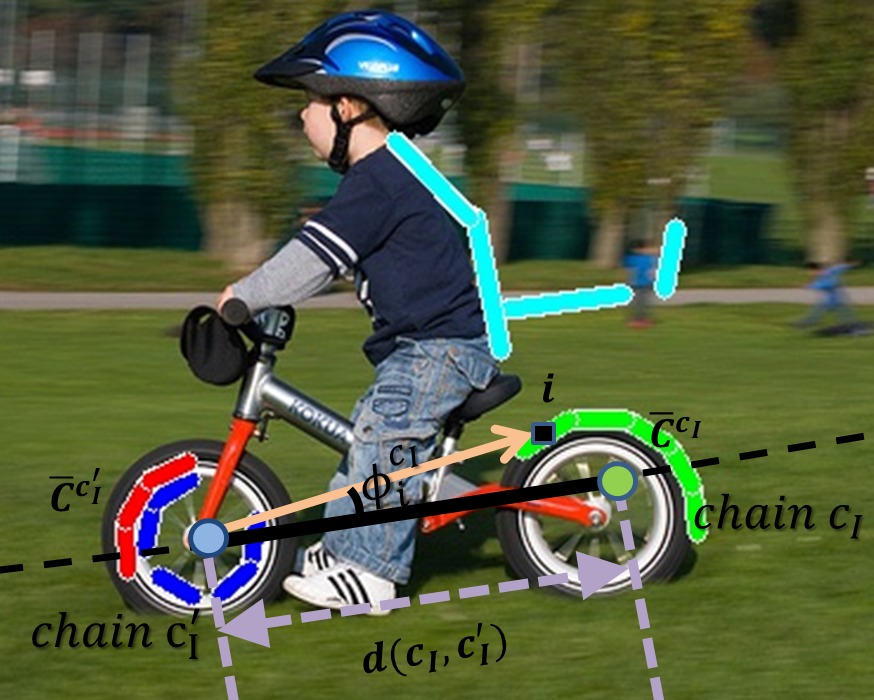} \label{b}}
\end{center}
\caption{Pairwise geometric consistency of the matched portions of a chain pair $\mathbf{p}=(C_S,C_I)$ with respect to $\mathbf{p^\prime}=(C^\prime_S, C^\prime_I)$ uses (i) the distances $d(C_S, C_S^\prime)$ and $d(C_I, C_I^\prime)$ between their centroids ($\overline{C}$) and (ii) the difference of angles  $\left|\phi^{C_S}_{i} - \phi^{C_I}_{i}\right|$.}
\label{fig:inter_chain_cycle}
\end{figure*}

The geometric consistency of the matched portions of a pair of chains $\mathbf{p}=(m(C_S),\allowbreak m(C_I))$ with respect to that of another chain pair $\mathbf{p^\prime}=(m(C^\prime_S), m(C^\prime_I))$, where $C_S$ and $C^\prime_S$ are the sketch chains and $C_I$, $C^\prime_I$ are the image chains, is measured based on two factors: a) \emph{distance-consistency} $G_d(\mathbf{p}, \mathbf{p^\prime})$ and b) \emph{angular-consistency} $G_a(\mathbf{p}, \mathbf{p^\prime})$.
The centroids of the matched chain portions can be obtained in a manner that is relatively robust to the presence of noise. Therefore, $G_d(\mathbf{p}, \mathbf{p^\prime})$ is defined in terms of the closeness of the distances between the chain centroids $d(m(C_S), m(C_S^\prime))$ in the sketch and $d(m(C_I), m(C_I^\prime))$ in the database image (Fig.~\ref{fig:inter_chain_cycle}). These distances are normalized by the total length of the matched portions of the corresponding chains in order to achieve scale insensitivity:
\begin{eqnarray}
%&G_{d}(\mathbf{p}, \mathbf{p^\prime}) = \exp \left( \lambda_{c} \cdot \left(1 - 1\middle/ \Omega\left( \frac{d(m(C_S), m(C_S^\prime))}{L_{S}}, \frac{d(m(C_I), m(C_I^\prime))}{L_{I}} \right) \right) \right) 
&G_{d}(\mathbf{p}, \mathbf{p^\prime}) = \exp \left( \lambda_{c} \cdot \left(1 - \middle. \Omega\left( \frac{d(m(C_S), m(C_S^\prime))}{L_{S}}, \frac{d(m(C_I), m(C_I^\prime))}{L_{I}} \right) \right) \right) 
\label{spatialConsistency}
\end{eqnarray}
where, $L_{S}$=length $(m(C_S)) +$ length$(m(C_S^\prime))$, $L_{I}$=length $(m(C_I))$+ length $(m(C_I^\prime))$ , $\lambda_{c}$(=1) is a scalar constant and $\Omega$ is defined in Eqn.~\ref{segmentLengthRatio}.

 %--------------------------------------------------------------FIG : INTER_CHAIN Cycle------------------------------------------
%INTER-CHAIN CYCLE
%---------------------------------------------------------------------------------------------------------------------------

The next factor $G_a$ measures \emph{angular-consistency}. To achieve rotational invariance, the line joining the corresponding chain centers is considered as the reference axis and the relative angle difference at the $i^{th}$ joint is determined (Fig.~\ref{fig:inter_chain_cycle}). 
$G_{a}(\mathbf{p}, \mathbf{p^\prime})$ is defined using the average difference of such relative angles of all the individual matched joints in a chain:
\begin{equation}
G_{a}(\mathbf{p}, \mathbf{p^\prime}) = \exp \left( - \lambda_a \cdot \frac{1}{N^{J_{\mathbf{p}}}} \sum\limits_{i=1}^{N^{J_{\mathbf{p}}}} \left|\phi^{C_S}_{i} - \phi^{C_I}_{i}\right| \right) 
\label{segmentwiseAngularConsistency} 
\end{equation}
where $N^{J_{\mathbf{p}}}$ is the number of matched joints between $C_S$ and $C_I$ and $\lambda_{a}$(=2) is a scalar constant.
Since, both $G_d$ and $G_a$ should be high for consistent matching, we consider the pairwise geometric consistency $G(\mathbf{p}, \mathbf{p^\prime})$ as a product of the constituent factors:
$G(\mathbf{p}, \mathbf{p^\prime}) =  G_d(\mathbf{p}, \mathbf{p^\prime}) \cdot G_a(\mathbf{p}, \mathbf{p^\prime})$.

Erroneously matched chains are typically geometrically inconsistent with others and one may have both geometrically consistent and inconsistent pairs in a group of matched pairs between a sketch and an image. Therefore, the geometric consistency $GC(\mathbf{p})$ for a matched pair $\mathbf{p}$ is taken to be the maximum of $G(\mathbf{p}, \mathbf{p^\prime})$ with respect to all other matched pairs $\mathbf{p^\prime}$: $GC(\mathbf{p}) = \operatorname*{max}_{\substack{\mathbf{p^\prime}}} G(\mathbf{p}, \mathbf{p^\prime})$. The max operator allows us to neglect the falsely matched pairs while considering only the consistent matched pairs. 
Finally, the similarity score of a database image $I$ with respect to a sketch query $S$ is determined as:
\begin{equation}
Score(S, I) = \sum\limits_{\mathbf{p}\in \text{P}} GC(\mathbf{p}) \cdot CS(	\mathbf{p})
\label{finalImageScore}
\end{equation}
where $CS(\mathbf{p})$ is the \emph{Chain Score} for the chain pair $\mathbf{p}$ (Eqn.~\ref{eqn:finalChainScore}) and $\text{P}$ is the set of all matched pairs of chains between a sketch $S$ and an image $I$.
Since erroneously matched chains get very low score for consistency, effectively only the geometrically consistent chains are given weight for scoring an image. 
This score is used to determine the final ranking of the database images, which can be used for ranked display of such images. 

\begin{figure*}[t]
\centering
\includegraphics[width=1\linewidth]{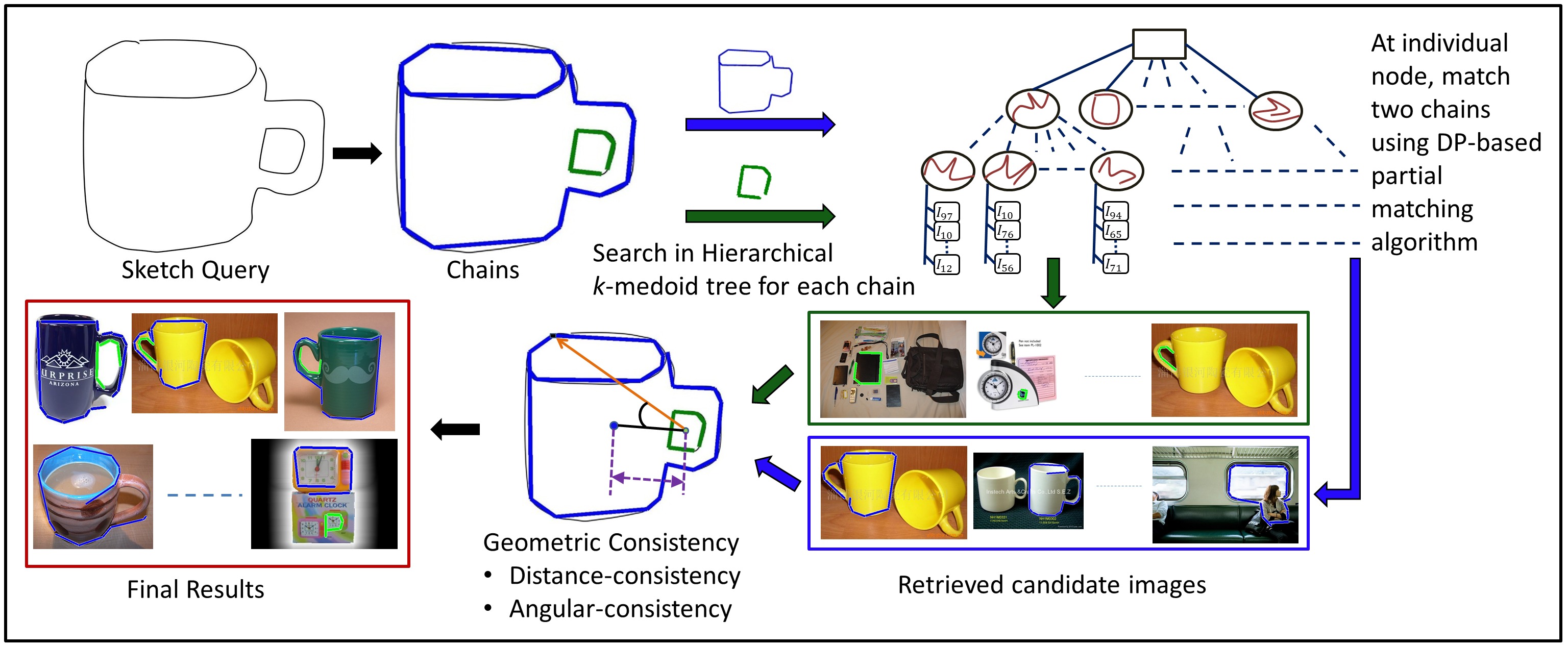} 
%\vspace{-7pt}
\caption{The entire Retrieval Framework}
%\vspace{-10pt}
\label{fig:searchFramework}
\end{figure*}

Fig.~\ref{fig:searchFramework} shows the complete retrieval framework. Results of experiments are considered next.

%-------------------------------------FIG: Search Framework ---------------------------------
%------------------------------------------------------------------------------------------------

%~~~~~~~~~~~~~~~~~~~~~~~~~~~~~~~~~~~~~~~~~~~~~~~~~~~~~~~~~~~~~~~~~~~~~~~~~~~~~~~~~~~End of Subsection: GEOMETRIC CONSISTENCY~~~~~~~~~~~~~~~~~~~~~~~~~~~~~~~~~~~~~~~~~~~~~~~~~~~~~~~~~~~

%-------------------------------------------------------------------------------------------END OF SECTION 5 : Image Retrieval  --------------------------------------------------------------------------------------

%-----------------------------------------------------------------------------------------BEGIN SECTION 6 : Experiments-------------------------------------------------------------------------
\section{Experiments}
\label{sec:experiments}

To evaluate the performance of our system, we created a database of $1.2$ million images, which contains $1$ million Flickr images taken from the MIRFLICKR-1M image collection~\citep{huiskes2008mir}. In addition, we included $0.2$ million images from the Imagenet~\citep{deng2009imagenet} database in order to have some common object images in our database. 
In the experiments, the hierarchical index for 1.2 million images is generated with a branching factor of $32$ and a maximum leaf node size of $100$, which leads to a maximum tree depth of 6. Using this tree, we obtain around $1500$ similar images for a given sketch, for which geometric consistency (Eqn.~\ref{finalImageScore}) is applied to finally rank the list of retrieved images from the chain tree.

The whole operation for a given sketch typically takes $1-5$ seconds on a single thread running on an Intel Core i7-3770 3.40GHz CPU. The running time typically depends on the number of chains in the sketch and most of the processing time is consumed by the geometric verification phase. 
However, this geometric consistency check can be trivially parallelized. \iffalse Therefore, the time consumed by this stage can be scaled down almost linearly with the number of cores. \fi

The hierarchical index for our dataset required only around $250$ MB of memory. %which could be easily stored on the RAM itself for fast access. 
For fast online access of database chain descriptors during geometric verification and ranking of retrievals, all the descriptors for $1.2$ million images are loaded a priori in the memory, which additionally required approximately $9$ GB of memory. Note that, the chain descriptors can be distributed across multiple CPUs if such geometric consistency check is performed parallely.  Furthermore, to make our approach work in a memory-constrained environment, for every sketch, only the descriptors corresponding to \texttildelow$1500$ selected images may be loaded each time in the memory at runtime although this may slow down the process somewhat due to online disk access. \iffalse Although the running time may increase slightly, this drastically reduces the additional memory requirement to only a few MBs. \fi
\iffalse Note that either only requried chain descriptors can be loaded in the memory on the fly while measuring the geometric consistency or all the chain descriptors may be loaded a priori for fast online access. We observed a memory footprint of approximately 9 GB while also loading the chain descriptors for all 1.2M images. \fi

%-----------------------------------------------------------FIG: Results---------------------------------------------
\begin{figure*}[h!]	
\begin{center}
\includegraphics[width=1.01\textwidth]{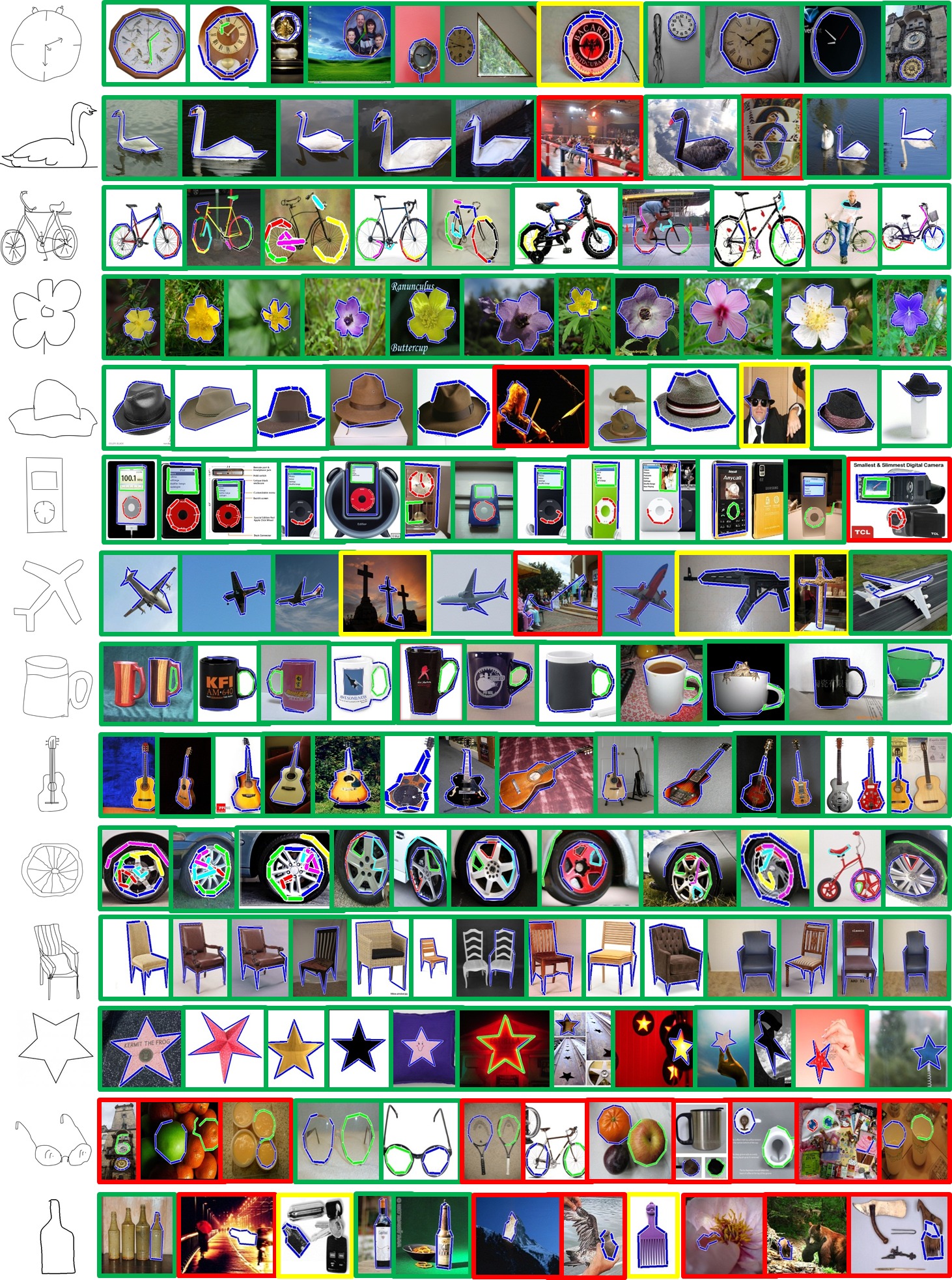}
\end{center}
\caption{Top retrieved images for 14 sketches from 1.2 million images. Retrieved images indicate similarity insensitivity and deformation handling of our approach. Chains are embedded on the retrieved images to illustrate the locations of the matchings. Multiple matched chains are shown using different colors. Correct, similar and false matches are illustrated by green, yellow and red boxes respectively (Best viewed in color).}
\label{fig:Result}
\end{figure*}
%-------------------------------------------------------------- -----------------------------------------------------

Visual results for $14$ sketches of different categories of varying complexity are shown in Fig.~\ref{fig:Result}. These clearly indicate insensitivity of our approach to similarity transforms (e.g positive retrievals of the swan sketch). Furthermore, due to our partial matching scheme, an object is retrieved even under a viewpoint change if a portion of the distinguishing shape structure of the object is matched (e.g $4^{th}$ image for swan). Global invariance to similarities as well as matching with flipped objects can be seen in the results for the sketches of swan and bi-cycle ($2^{nd}$ and $10^{th}$ retrieved image for swan, $2^{nd}$ and $9^{th}$ retrieved image for bi-cycle etc.). 
It can be easily observed that the performance of our approach depends on the complexity/distinctiveness of the shape structure. False matches (e.g cross for the sketch of airplane; two adjacent clocks/cups for the sketch of spectacle; keychain, brush for the sketch of bottle in Fig.~\ref{fig:Result}) typically occur due to some shape similarity between the sketch and an object in the image, the probability of which is higher when the sketch is simple and/or contains only one chain (e.g bottle).

\begin{figure*}[t]	
\begin{center}
\includegraphics[width=1\linewidth]{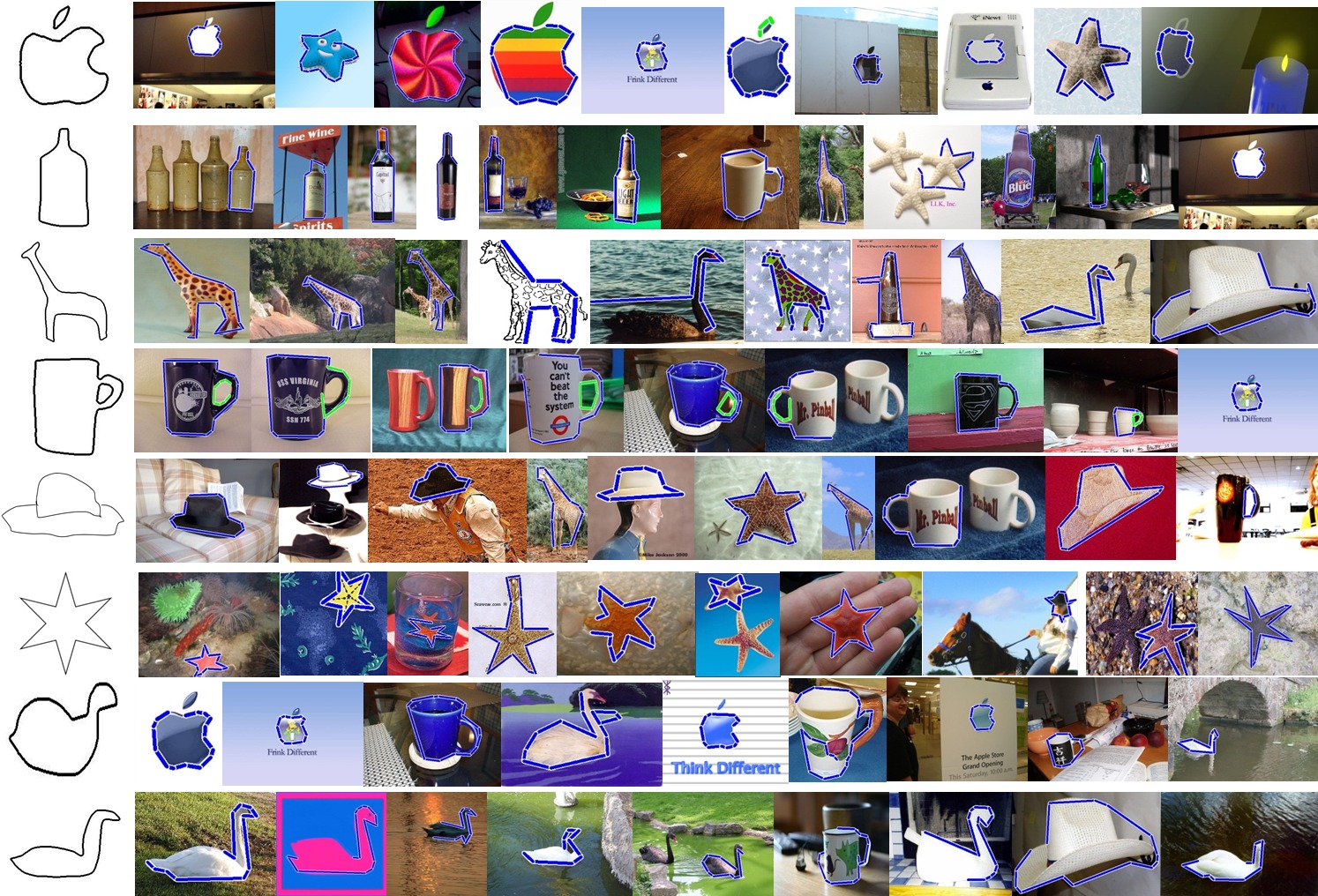}
\end{center}
%\vspace{-10pt}
\caption{Top retrievals from ETHZ extended shape dataset~\citep{schindler2008object} for few sketches.}
%\vspace{-7pt}
\label{fig:retrievalsInETHZ}
\end{figure*}

\iffalse The performance of our approach depends on the complexity/distinctiveness of the shape structure and it can be easily observed from \fi 
To understand the characteristics of the missed retrievals as well in a controlled dataset, we also tested our system on the ETHZ extended shape dataset consisting of $385$ images of $7$ different categories with significant scale, translation and rotation variations. Fig.~\ref{fig:retrievalsInETHZ} shows top retrieval results from the ETHZ extended shape dataset~\citep{schindler2008object} for a few sketches. It can be observed that the accuracy of retrieval heavily depends on the quality of the sketch. 
\iffalse 
Fig.~\ref{fig:retrievalsInETHZ} shows top retrieval results from ETHZ extended shape dataset~\cite{schindler2008object} for some of the sketches, whereas Fig.~\ref{fig:top_20_ethz} details the performance of our approach for different categories.
As discussed earlier, the performance of our approach depends on the complexity/distinctiveness of the shape structure and the quality of the user sketch. \fi
In Fig.~\ref{fig:retrievalsInETHZ}, a significant portion of the first swan ($7^{th}$ sketch) is circular and thus it matches to locally circular shapes. However, for a relatively better sketch of swan ($8^{th}$ sketch), the number of positive retrievals is higher.
Sometimes, the matched portions of two different shapes appear to be globally similar, which again leads to false positives at a few top positions (e.g matched portion of giraffe and mug for the sketch of hat in Fig.~\ref{fig:retrievalsInETHZ}).

%----------------------------------------------------------FIG : Retrievals in ETHZ-----------------------

%---------------------------------------------End of Figure: Retrievals in ETHZ-----------------------------------------------------

\begin{comment}
In general, our approach performs worse when the object shape is very simple, i.e. the amount of distinctiveness in the shape structure is less. In Fig. X, the sketch of applelogo, due to simplicity in the shape structure, matches to lot of locally circular objects. 
Furthermore, the retireval accuracy highly depends on the complexity of the drawn sketch. In Fig. X, it can be observed that significant portion of the Xth swan is circular and thus it matches to locally circular shapes. 
\end{comment}

\begin{figure*}[t]	
\begin{center}
\subfloat[][]{\includegraphics[width=0.47\linewidth]{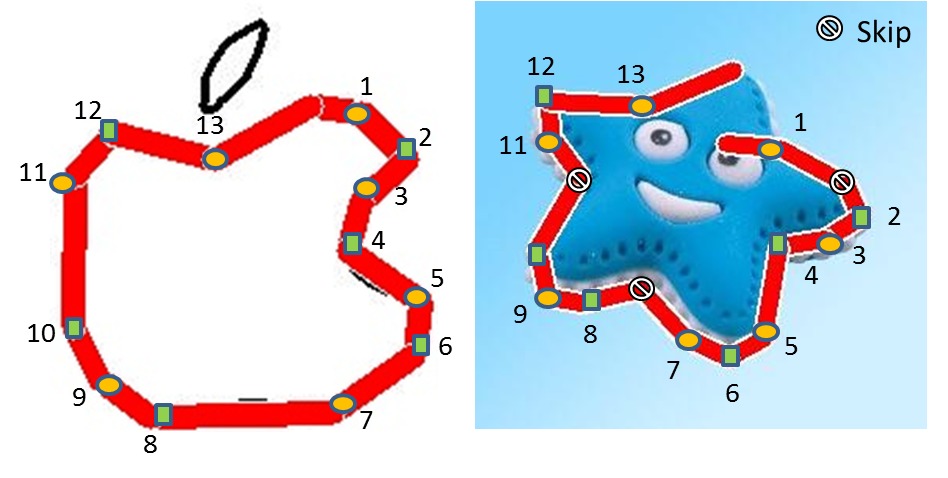}\label{a}}\qquad
\subfloat[][]{\includegraphics[width=0.47\linewidth]{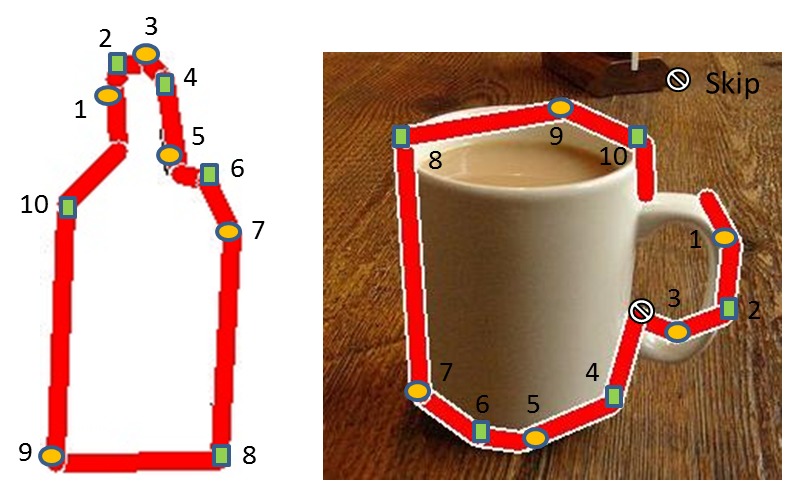} \label{b}}
\end{center}
\caption{Skipping of important joints and allowing local deformations lead to false positive retrievals. Matched joints are numbered same in both the sketch and the image.}
\label{fig:FailureCaseOfEntireMethod}
\end{figure*}

Even when the object shape is well captured in both the sketch and the image, sometimes our approach retrieves an incorrect object if the distinctive object boundary portion(s) get skipped. Fig.~\ref{fig:FailureCaseOfEntireMethod}(a) illustrates such a situation where skipping of very important object boundary portions in the image of ``star'' leads to a wrong retrieval. Furthermore, the matched portions of corresponding chains are globally similar. Therefore, even global angular consistency check fails to identify the false retrieval in this case. 
This case cannot be easily addressed if we are to allow skips to handle noise in boundary extraction. 
Note that, for a sketch of ``star'', however, an image of apple does not get a high matching score because of asymmetricity in our chain matching criteria where we assume that the sketch chains are much less noisy than the image chains and hence skips are more heavily penalized in the sketch chains. 
\iffalse 
We assume that sketch chains are not noisy and it contains only object shape information. Therefore, to retrieve the intended object, we penalize skips in sketch chain heavily. In contrast, database chains are typically noisy and sometimes non-object boundary portion becomes part of the chain even within the object boundary portion. Thus, with a goal to match the object portion, we allow comparatively more skips in database chains. 
Therefore, in case of star(sketch)-to-apple(image) matching, the chances that the distinctive portions of a star (i.e. the successive turning points) getting skipped is less. \fi
This also explains the reason behind very good retrievals for the objects whose shape is (almost) unique. 
Fig.~\ref{fig:FailureCaseOfEntireMethod}(b) also shows an example where allowing local deformation and skipping an informative joint lead to a false positive. 
% However, sophisticated and efficient local matching algorithms that simultaneously enforces global similarity can be used to improve the results further and is considered as a future work.

\iffalse Further, It can be observed that for highly deformable objects, viz. giraffe, the performance is quite poor. Sofisticated algorithms can be used for handling high amount of deformation and is considered as a future work. \fi

%--------------------------------------------------------------FIG : Failure case of Entire Method ------------------------------------------

%---------------------------------------------------End of Figure: Failure case of entire method------------------------------------------------------------------------

Quantitative measurement of the performance of a large scale retrieval system is not easy due to the difficulty in obtaining ground truth data, which is currently unavailable for a web-scale dataset. Some common metrics to measure retrieval performances (F-measure, Mean Average Precision~\citep{manning2008introduction} etc.) use recall which is impossible to compute without a full annotation of the dataset. Therefore, to evaluate the performance of our approach quantitatively, we use the Precision-at-$K$ measure for different rank levels ($K$) for the retrieval results~\citep{rubner2000earth}. This is an acceptable measure since an end-user of a large scale Image Retrieval system typically cares only about the top results which must be good. 

First, we separately evaluate the major components of our method to understand their effects. Then, we discuss the retrieval performance of the proposed algorithm in comparison to prior work on our large dataset. To study the properties of our algorithm in detail, the proposed algorithm is also evaluated on the ETHZ shape dataset~\citep{ferrari2006object}, on which it is possible to compute the recall as well. Finally, using these evaluations, we discuss the strengths and weaknesses of our approach.

\subsection{Evaluation of major components}
\label{sec:evaluationOfMajorComponents}
User sketches are highly subjective~\citep{eitz2012humans} and the retrieval performance depends on the quality of the user sketch. Therefore, to obtain a robust estimate of the performance, the system must be tested using a diverse set of user sketches of varying complexity. To this end, we use a dataset (\cite{shreyashiCVIU}) of $50$ user-drawn sketches consisting of $10$ sketches for each of the five shape categories in the ETHZ shape dataset~\citep{ferrari2006object}, viz. applelogo, bottle, giraffe, mug and swan. This dataset provides a wide variety in terms of the quality of the sketch and is therefore an appropriate choice for evaluation purposes. For testing these sketches, we add the images of the ETHZ shape dataset~\citep{ferrari2006object} to our dataset of $1.2$ million images. Finally, for each of these sketches, we determine the number of correct matches in the top $50$ retrievals, where such counting has to be done manually as we do not possess any prior categorical information of the dataset images.

Two chain extraction strategies are used in our work to capture the object shape information: a) overlapping long chains from contour segment networks and b) the boundaries of the segmented object proposals. 
To understand the relative benefits of these, we first measure the retrieval performance considering only a single method of chain extraction for all the database images at a time. 
Table~\ref{table:majorComponentsEvaluationTable} shows the retrieval scores for different object categories. 
Allowing overlaps while extracting long chains from contour segment networks (OC+GC) compared to using only non-overlapping chains (NOC+GC)~\citep{parui2014similarity} increases the chance of covering the entire object boundary and hence gives superior results. \iffalse Therefore its retrieval performance is better than using only non-overlapping chains (NOC+GC)~\citep{parui2014similarity}. \fi When the chains are extracted using segmented object proposals (GOP+GC)~\citep{krahenbuhl2014geodesic}, improved results can be observed for a few categories. For some objects, segmentation produces better result and if the entire object is covered in a single segmented proposal, then an accurate chain corresponding to the object boundary can be extracted. Therefore, for such categories, viz. appelogo, swan etc., one obtains a good accuracy when chains extracted using segmented proposals are used. 
\begin{comment}
Considerable performance improvement is observed for few categories when the chains extracted using segmented object proposals~\cite{krahenbuhl2014geodesic} are used instead (GOP+GC). This is primarily because very accurate and precise object boundary can be extracted \iffalse for comparatively simpler shapes \fi using these proposals. 
Typically simpler shape categories have lesser distinctive portions in their shape structure.
\iffalse Some object categories have comparatively lesser distinctive portions in their shape structure. \fi
For example, there is a subtle difference between any circular object and apple/applelogo. Therefore, for these categories, boundary representation without any deviation due to noise is essential. 
If the entire object is covered in a segmented proposal, then an accurate chain corresponding to the object boundary can be extracted. Therefore, for few categories,viz. applelogo, swan retrieval accuracy increases when chains extracted using segmented proposals are used. 
\end{comment}
However, for other categories, the top proposals cover only small and/or ambiguous portions of the object(s). In such cases (giraffe, bottle), a better retrieval score is obtained when chains are extracted from the contour segment networks.
To utilize the benefits of both these mechanisms, chains extracted from these two methods are combined for all the database images (OC+GOP+GC) and significant improvement on retrieval performance can be observed as compared to the different ideas considered in isolation or compared to only using the non-overlapping chains~\citep{parui2014similarity}.
\iffalse In case of apple/applelogo, proposals are typically better because smaller regions tend to get higher score among different proposals~\cite{gop}. \fi

%--------- TABLE for Major Component Evaluation on shreyashi's dataset -----------
\begin{table}[t]
\centering
    \begin{tabular}{|l|l|l|l|l|l|}
    \hline
    \textbf{Method}    & \textbf{Applelogo} & \textbf{Bottle} & \textbf{Giraffe} & \textbf{Mug}  & \textbf{Swan}  \\ \hline
    NOC+GC    & $21.4\pm2.6$      & $10.2\pm2.4$   & $7.8\pm2.7$     & $84.4\pm3.2$ & $50.9\pm2.8$ \\ \hline
    OC+GC     & $22.2\pm2.4$        & $13.8\pm3.6$   & $12.2\pm4.4$    & $90.6\pm1.8$ & $58.4\pm3.4$ \\ \hline
    GOP+GC    & $26.8\pm3.5$      & $13\pm2.6$   & $7.2\pm2.3$       & $91\pm1.9$   & $64\pm2.9$ \\ \hline
    OC+GOP    & $25.6\pm1.8$      & $14.6\pm3.2$   & $6.2\pm1.4$       & $70.4\pm2.9$   & $64.7\pm2.1$ \\ \hline
    OC+GOP+GC & $27.6\pm2.83$      & $14.6\pm3.2$   & $12.4\pm4.3$    & $93.8\pm1.5$ & $65.1\pm2.2$ \\ \hline
   \begin{tabular}{@{}c@{}}OC+GOP+GC \\ (ETHZ Models)\end{tabular} & \multicolumn{1}{|c|}{$54$} & \multicolumn{1}{|c|}{$16$} & \multicolumn{1}{|c|}{$20$} & \multicolumn{1}{|c|}{$94$} & \multicolumn{1}{|c|}{$80$} \\ \hline
    \end{tabular}
    \caption {Percentage of true positive images in top 50 retrievals with the corresponding standard deviation using different chain extraction mechanisms. NOC: Non-Overlapping Chains~\citep{parui2014similarity}, OC: Chains with Overlap allowed, GOP: Chains extracted using Geodesic Object Proposal~\citep{krahenbuhl2014geodesic}. OC+GOP uses chains extracted from both methods. GC indicates performance with geometric verification. Last row details the performance when ETHZ Models~\citep{ferrari2006object} are used.}
    \label{table:majorComponentsEvaluationTable}
\end{table}

%--------- END of TABLE for Major Component Evaluation on shreyashi's dataset -----------

Table~\ref{table:majorComponentsEvaluationTable} also shows the importance of considering geometric consistency of the matched chains (OC+GOP+GC vs. OC+GOP). 
\iffalse Recall that individual chains are matched in a scale, translation and rotation insensitive way. 
Therefore considering geometric consistency between the matched portion helps us to reduce the score of falsely matched chains (Fig.~\ref{fig:indexTreeAndReqOfGeometricConsistency}(b)) and thereby eliminate them. \fi
Although significant performance gain is observed after applying geometric consistency, it is not possible to apply this step when only one chain is matched between a sketch and a database image. 
\iffalse In that case, the chain matching score (Eqn.~\ref{eqn:finalChainScore}) is considered as the final matching score of the image for a given sketch. \fi
Due to this, for some of the categories (viz. bottle, swan etc.) in Table~\ref{table:majorComponentsEvaluationTable}, applying geometric consistency does not make much difference in the retrieval score.

Object categories vary in the complexity and uniqueness of their shape information. 
\iffalse Therefore, a wide variation in the retrieval accuracy can be observed in Table~\ref{table:majorComponentsEvaluationTable} for different categories. \fi
For highly deformable objects,viz. giraffe, the performance is poor since our approach can only handle a similarity transform.
Furthermore, there is a considerable amount of texture variation and background clutter for the giraffe images in the ETHZ shape dataset~\citep{ferrari2006object}, making
it hard to extract good chains for this category.
\iffalse Thus the extracted chains from either method fails to cover substantial and/or important portions of the object boundary, which also attributes to low retrieval score for the sketches of giraffe. \fi
For simpler shapes, viz. apple, bottle etc., many false positives get a good matching score as these shapes are relatively simpler and thus easy to match. 
Typically, our approach performs well for object categories with a distinctive shape and where the chains can be extracted easily. 
The influence of the sketch quality is also evident from the standard deviation in the retrieval accuracies. 
Table~\ref{table:majorComponentsEvaluationTable} also lists the performance of our approach for different categories using the fairly good quality ETHZ dataset model sketches~\citep{ferrari2006object}, for which much better performance was obtained.
\iffalse
Table~\ref{table:majorComponentsEvaluationTable} also lists the performance of our approach for different categories when ETHZ model shapes~\citep{ferrari2006object} are used. In this case, the performance is better than the average performance on user-drawn sketches (\cite{shreyashiCVIU}), which is expected since those sketches are computer generated and are therefore much cleaner. \fi

Next, we show much more comprehensive retrieval results of the proposed algorithm and compare them to prior work.

\subsection{Comparisons with Prior Work}
\label{sec:resultsAndComparisonWithPriorWork}
\iffalse First, both qualitative and quantitative comparison is performed on our large dataset. Then, to understand the behavior of our algorithm better, results are shown on a standard shape dataset (ETHZ shape dataset~\citep{ferrari2006object}). A comparative study is further performed on this small dataset and challenging cases of our approach are highlighted. \fi

\subsubsection{Million Image Dataset}

To evaluate our system for large scale retrieval on different object categories, we asked $5$ random subjects to draw sketches for a variety of objects on a touch-based tablet and collected $75$ sketches. These sketches, along with $100$ sketches from a crowd-sourced sketch database~\citep{eitz2012humans}, containing $24$ different categories in total, are used for retrieval. 
Non-availability of a public implementation of any prior work makes it difficult to have a comparative study with prior work.
Even though a Windows phone App (\citeauthor{sketchmatch}) based on~\citep{cao2011edgel} is available, the database is not available to make a fair comparison to other algorithms. Hence, we re-implemented this algorithm~\citep{cao2011edgel} (EI) as well as another by~\cite{eitz2009descriptor} (TENSOR) and tested their algorithms on our database for the purpose of comparison.~\cite{zhou2012sketch} did not provide complete implementation details in their publication and it is not trivial to make the method proposed by~\cite{riemenschneider2011image} run efficiently on a very large database. Furthermore,~\cite{riemenschneider2011image} did not show any result on a large scale dataset and~\cite{zhou2012sketch} showed results only for 3 sketches. Hence, these methods were not compared against. 

\begin{figure*}[t]	
\begin{center}
\includegraphics[width=0.94\linewidth]{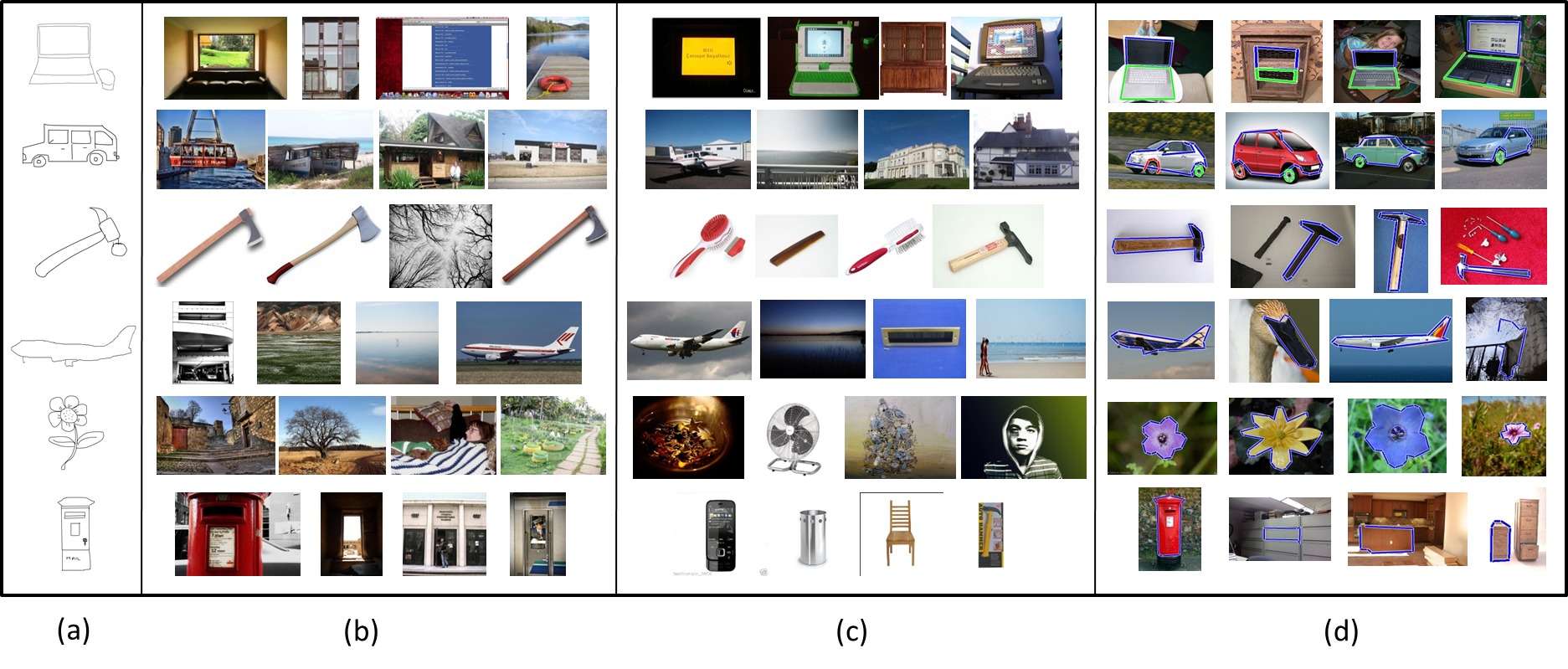}
\end{center}
%\vspace{-10pt}
\caption{Top 4 results by (b)~\cite{eitz2009descriptor}, (c)~\cite{cao2011edgel} and (d) our system on a 1.2 million image dataset for some sample sketches (a).}
%\vspace{-7pt}
\label{fig:comparison}
\end{figure*}

%--------- TABLE for Large Scale Retrieval -----------
\begin{table}[t]
\scriptsize
\caption{Precision (expressed as Percentage of true positives) at different ranks for $175$ retrieval tasks in $24$ categories on a dataset of $1.2$ million images. B: Best, W: Worst, A: Average performances are computed among sketches for each category and then averaged. }
\centering
\tabcolsep=0.05cm
\begin{tabular}{ |l |  c c c  |  c c c  |  c c c  |  c c c  |  c c c  |  c c c |}
      \hline 
      \multicolumn{1}{c}{} & & & \\[\dimexpr-\normalbaselineskip-\arrayrulewidth]% Correct for mis-alignment
      {\bf Method} & \multicolumn{3}{c}{\bf Top 5} & \multicolumn{3}{c|}{\bf Top 10} & \multicolumn{3}{c|}{\bf Top 25} & \multicolumn{3}{c|}{\bf Top 50} & \multicolumn{3}{c|}{\bf Top 100} & \multicolumn{3}{c|}{\bf Top 250} \\
          & B & W & A & B & W & A & B & W & A & B & W & A & B & W & A & B & W & A \\
    \hline
    \begin{tabular}{@{}l@{}}{\bf TENSOR} \\ ~\citep{eitz2009descriptor}\end{tabular} & 30.8 & 7.5 & 14.7 & 30 & 7.1 & 13.7 & 24.8 & 7 & 12.9 & 20.8 & 7 & 12.3 & 16.5 & 5.8 & 10.2 & 9.4 & 3 & 5.7\\ \hline
    \begin{tabular}{@{}l@{}}{\bf EI} \\ ~\citep{cao2011edgel}\end{tabular}  & 36.7 & 20.8 & 23.4 & 34.2 & 17.9 & 21.5 & 30 & 15.3 & 19.5 & 27 & 13.8 & 17.5 & 22.2 & 11.2 & 14.8 & 15.7 & 7.8 & 10.5  \\ \hline
    \begin{tabular}{@{}l@{}}{\bf NOC+GC} \\~\citep{parui2014similarity}\end{tabular}  & 80.8 & 42.5 & 60.8 & 72.5 & 38.3 & 53.6 & 54.7 & 29.3 & 39.5 & 40.3 & 20.7 & 28.5 & 31.8 & 16.3 & 22.2 & 23 & 12.5 & 16.5  \\ \hline
    {\bf OC+GOP+GC} (Ours)  & \textbf{86.5} & \textbf{43.7} & \textbf{65.5} & \textbf{78.1} & \textbf{38.4} & \textbf{58.2} & \textbf{62.4} & \textbf{30.5} & \textbf{44.6} & \textbf{51.7} & \textbf{23.8} & \textbf{34.6} & \textbf{46.7} & \textbf{21.5} & \textbf{30.8} & \textbf{34.1} & \textbf{15} & \textbf{21.7} \\
%     OC+GOP+GC  & \textbf{86.5} & \textbf{43.7} & \textbf{66.1} & \textbf{78.1} & \textbf{38.4} & \textbf{58.2} & \textbf{62.4} & \textbf{30.5} & \textbf{44.6} & \textbf{51.7} & \textbf{23.8} & \textbf{34.6} & \textbf{46.7} & \textbf{21.5} & \textbf{30.8} & \textbf{34.1} & \textbf{15} & \textbf{21.7} \\
    \hline
    \end{tabular}  
\label{table:precision_large_scale}
\end{table}
%-- Table END for Large scale Retrieval--------------------
Table~\ref{table:precision_large_scale} shows the performances of our algorithm in comparison with TENSOR~\citep{eitz2009descriptor} and EI~\citep{cao2011edgel} at different retrieval levels.
First, the precision is computed for all sketches of a given object category and then the best, worst and average retrieval scores of different categories are averaged over all $24$ categories. The significant deviation between the best and the worst retrieval performances indicate the diversity in the quality of the user sketches and the system response to it. It can be observed from Table~\ref{table:precision_large_scale} that our method significantly outperforms the other two methods on this large dataset. Both TENSOR~\citep{eitz2009descriptor} and EI~\citep{cao2011edgel} consider edge matchings approximately at the same location in an image as that of the sketch and therefore, the retrieved images from their system contain the sketched shape only at the same position, scale and orientation while images containing the 
sketched object at a different scale, orientation and/or position are missed leading to false retrievals in top few matches (Fig.~\ref{fig:comparison}).
Similar performance was observed by us on the \emph{Sketch Match} app (\citeauthor{sketchmatch}), although a direct comparison with it is inappropriate since the databases are different.
\iffalse To evaluate the advantage of the geometry-based verification step, we also show the retrieval performance with and without this step and it can be observed that the geometric consistency check improves our results substantially. \fi
It can also be observed that using two types of chain extraction strategy and considering global angular consistency improved the performance compared to only using the non-overlapping chains~\citep{parui2014similarity}.
Note that, due to the non-availability of a fully annotated dataset of a million images, it is extremely hard to use an automated parameter learning algorithm. Hence, parameters are chosen empirically by trying out a few variations. Better parameter learning/tuning could possibly improve the results further.

%----------------------------------------------------------FIG : Comparison-----------------------
%Comparison
%--------------------------------------------------------------------------------------------------

\subsubsection{ETHZ Dataset}
\label{sec:ethzDataset}

\begin{comment}
%-----------------TABLE and Figure for ETHZ
\noindent\hrulefill\par
\noindent\makebox[\linewidth][c]{%
 \begin{minipage}[b]{0.44\textwidth}
      \centering
      \scriptsize \small
      \captionof{table}{Comparison of \% of true positive retrievals in top 20 using our $63$ sketches and ETHZ models~\cite{ferrari2010images} on ETHZ dataset~\cite{ferrari2010images}}
      \tabcolsep=0.06cm
     \begin{tabular}{ |l |  c c c  |  c | }
      \hline
      \multicolumn{1}{c}{} & & & \\[\dimexpr-\normalbaselineskip-\arrayrulewidth]% Correct for mis-alignment
      Method & \multicolumn{3}{c}{Our Sketches} & ETHZ \\
	      & Best & Worst & Avg & Models~\cite{ferrari2006object}\\
    \hline
    TENSOR~\cite{eitz2009descriptor} & 17 & 10 & 13.5 & 13.6 \\ \hline
    EI~\cite{cao2011edgel} & 46 & 6 & 26.7 & 27.9\\ \hline
    NOC+GC~\cite{parui2014similarity} & 60 & 28 & 42.9 & 49.3 \\ \hline
    OC+GOP+GC & \textbf{76} & \textbf{37} & \textbf{56.3} & \textbf{76.4}  \\ 
    \hline
    \end{tabular}
    %\vspace{12pt}	
    \label{table:precision_ethz}
\end{minipage}
\qquad
\begin{minipage}[b]{0.49\textwidth}
   \centering
    \includegraphics[width=1\textwidth]{comparison/top20_ethz_self_CROPPED}
    %\vspace{-10pt}
    \captionof{figure}{Retrieval performance of the proposed algorithm for different categories of the ETHZ shape dataset~\cite{ferrari2010images}}
    \label{fig:top_20_ethz}
\end{minipage}
}
\noindent\makebox[\linewidth]{\rule{\linewidth}{0.2pt}}
%-----------_End of Table and Figure for ETHZ-------------------------------------------------
\end{comment}

To provide comparisons on a standard dataset and study the recall characteristics which is difficult for a large dataset, we tested our system on the ETHZ shape dataset~\citep{ferrari2006object}. \iffalse consisting of 283 images of 5 different categories with significant scale, translation and rotation variation. \fi
We used the user-drawn sketches of~\cite{shreyashiCVIU} for evaluation purposes.
Although standard sketch-to-image matching algorithms for Object Detection that perform time consuming online processing certainly perform better than our approach on this small dataset, such comparison would be unfair since the objectives are different. Hence, we compare only against TENSOR~\citep{eitz2009descriptor} and EI~\citep{cao2011edgel}. In this dataset, we measure the percentage of positive retrievals in top $20$ retrieved results which also gives an idea of recall of various approaches since the number of true positives is fixed.
Table~\ref{table:precision_ethz} shows the best, worst and average performance for the different sketches in a category (as for the previous dataset). It can be seen that our method performs much better than other methods on this dataset as well. 
The retrieval performance on ETHZ models~\citep{ferrari2006object} further indicates substantial advantage of using very good sketches.
\iffalse The performance on ETHZ models~\cite{ferrari2006object} is better than the average performance, which is expected since those sketches are computer generated and are therefore cleaner. \fi

%--------- TABLE for ETHZ Comparison with Prior Work -----------
\begin{table}[t]
\scriptsize
\caption{Comparison of Percentage of true positive retrievals in top $20$ using $50$ sketches of~\cite{shreyashiCVIU} and ETHZ models~\citep{ferrari2006object} on ETHZ dataset~\citep{ferrari2006object}.}
\centering
\begin{tabular}{ |l |  c c c  |  c | }
    \hline
    \multicolumn{1}{c}{} & & & \\[\dimexpr-\normalbaselineskip-\arrayrulewidth]% Correct for mis-alignment
    {\bf Method} & \multicolumn{3}{c}{{\bf Our Sketches}} & {\bf ETHZ Models}\\
	    & Best & Worst & Avg & \citep{ferrari2006object}\\
  \hline
  {\bf TENSOR}~\citep{eitz2009descriptor} & 17 & 10 & 13.5 & 13.6 \\ \hline
  {\bf EI}~\citep{cao2011edgel} & 46 & 6 & 26.7 & 27.9\\ \hline
  {\bf NOC+GC}~\citep{parui2014similarity} & 60 & 28 & 42.9 & 49.3 \\ \hline
  {\bf OC+GOP+GC} (Ours) & \textbf{76} & \textbf{37} & \textbf{56.3} & \textbf{76.4}  \\ 
  \hline
  \end{tabular}
\label{table:precision_ethz}
\end{table}
%-- Table END for ETHZ Comparison with Prior Work--------------------

%---------------------------------------------------------------------------------------END of SECTION 5 : Experiments-------------------------------------------------------------------------

%-----------------------------------------------------------------------------BEGIN CONCLUSION-------------------------------------------------------------------------------------------------
\section{Conclusions}
We have proposed an efficient image retrieval approach via hand-drawn sketches for large datasets. To the best of our knowledge, this is the first major work in the field of large scale sketch-based image retrieval that handles rotation, translation, scale and small variations of the object shape even for a dataset consisting of millions of images. This is accomplished by representing the images using chains of contour segments that have a high probability of containing the object boundary. \iffalse using two complimentary methods. We have argued the flexibility of the system for adapting different boundary representation mechanisms.\fi A similarity-invariant variable length descriptor is proposed that is used to partially match two chains in a hierarchical indexing framework. We also proposed a geometric verification scheme for improving the search accuracy. Experimental results shown on different datasets clearly indicate the benefits of our approach compared to the existing methods. 

Due to similarity-invariance of our approach as compared to other relevant work, our method could be used to efficiently search in large natural image databases, which typically have a lot of variations. The proposed method could also open the window for efficiently searching in constrained image databases, viz. personal photo albums, which typically do not contain any tag/text information. Furthermore, our method could be augmented by other techniques such as text for tagging images in an offline fashion. \iffalse or for improving online results.\fi

One major issue of our approach is the difficulty in extracting ``good'' representative chains in the presence of considerable background clutter. Newer and better boundary extraction mechanisms can be easily adapted to our framework for improving the quality of the chains.
Furthermore, the proposed sketch-to-image matching approach is only similarity-invariant and also fails to handle substantial deformation and major viewpoint changes. A sophisticated affine or even projective-invariant matching mechanism could possibly help retrieve images with such variations as well and can be considered in future work.

\bibliographystyle{spbasic_Sarthak} 
\bibliography{bib_ijcv_5.bib}   % name your BibTeX data base

\end{document}